%% file: main.tex
\documentclass{article}

\PassOptionsToPackage{numbers}{natbib}

\usepackage[preprint]{neurips_2025}

\input{packages}

\title{\acronym: A multi-sensor dataset for 6D pose estimation of chairs in industrial settings}

\author{%
    Mattia Nardon\thanks{Corresponding author.}\\
    FBK-TeV\\
    \texttt{mnardon@fbk.eu}\\
    \And
    Mikel Mujika Agirre\\
    Ikerlan\\
    \texttt{mmujika@ikerlan.es}\\
    \And
    Ander González Tomé\\
    Ikerlan\\
    \texttt{ander.gonzalez@ikerlan.es}\\
    \And
    Daniel Sedano Algarabel\\
    Ikerlan\\
    \texttt{dsedano@ikerlan.es}\\
    \And
    Josep Rueda Collell\\
    Ikerlan\\ 
    \texttt{rueda\_999@hotmail.com}\\ 
    \And
    Ana Paola Caro\\
    Andreu World\\
    \texttt{ap.caro@andreuest.com}\\
    \And
    Andrea Caraffa\\
    FBK-TeV\\
    \texttt{acaraffa@fbk.eu}\\
    \And
    Fabio Poiesi\\
    FBK-TeV\\
    \texttt{poiesi@fbk.eu}\\
    \And
    Paul Ian Chippendale\\
    FBK-TeV\\
    \texttt{chippendale@fbk.eu}\\
    \And
    Davide Boscaini\\
    FBK-TeV\\
    \texttt{dboscaini@fbk.eu}\\
}

\begin{document}

\input{commands}

\maketitle

\input{sections/0_abstract}
\input{sections/1_intro}
\input{sections/2_related}

\input{sections/3_dataset}
\input{sections/4_results}
\input{sections/5_conclusion}


\bibliographystyle{plainnat}  
\bibliography{main}

\end{document}

%% file: packages.tex
\usepackage[utf8]{inputenc} 
\usepackage[T1]{fontenc}    

\usepackage{microtype}
\usepackage{balance}
\usepackage{xspace}

\usepackage{amsmath,amssymb,amsfonts}

\usepackage{graphicx}
\usepackage[dvipsnames]{xcolor}
\usepackage{overpic}

\usepackage{booktabs}
\usepackage{multirow,multicol}
\usepackage{colortbl}

\usepackage{amsfonts}
\usepackage{pifont}
\usepackage{nicefrac} 

\usepackage{url}
\usepackage[pagebackref,breaklinks,colorlinks]{hyperref}

%% file: commands.tex
\newcommand{\warning}[1]{\textbf{\color{red!90}{#1}}}

\definecolor{greencamera}{RGB}{86, 196, 127}
\definecolor{purplecamera}{RGB}{172,106,181}
\definecolor{cyancamera}{RGB}{90,212,212}
\definecolor{myazure}{rgb}{0.8509,0.8980,0.9412}

\newcommand{\cmark}{\ding{51}}
\newcommand{\xmark}{\ding{55}}

\newcommand{\acronym}{CHIP\xspace}
\newcommand{\zed}{ZED 2i\xspace}
\newcommand{\rsd}{RealSense D435\xspace}
\newcommand{\rsl}{RealSense L515\xspace}

%% file: sections/0_abstract.tex
\begin{abstract}
Accurate 6D pose estimation of complex objects in 3D environments is essential for effective robotic manipulation.
Yet, existing benchmarks fall short in evaluating 6D pose estimation methods under realistic industrial conditions, as most datasets focus on household objects in domestic settings, while the few available industrial datasets are limited to artificial setups with objects placed on tables.
To bridge this gap, we introduce \acronym, the first dataset designed for 6D pose estimation of chairs manipulated by a robotic arm in a real-world industrial environment.
\acronym includes seven distinct chairs captured using three different RGBD sensing technologies and presents unique challenges, such as distractor objects with fine-grained differences and severe occlusions caused by the robotic arm and human operators.
\acronym comprises 77,811 RGBD images annotated with ground-truth 6D poses automatically derived from the robot's kinematics, averaging 11,115 annotations per chair.
We benchmark \acronym using three zero-shot 6D pose estimation methods, assessing performance across different sensor types, localization priors, and occlusion levels.
Results show substantial room for improvement, highlighting the unique challenges posed by the dataset.
\acronym will be publicly released.
\end{abstract}

%% file: sections/1_intro.tex
\section{Introduction}\label{sec:intro}

\input{figures/teaser/teaser}

Object 6D pose estimation involves determining the rotation and translation of objects in environments from sensory data.
It is essential for industrial automation, such as robotic painting, where predefined recipes are followed. 
Pose estimation enables a robotic arm to accurately position itself to consistently paint all parts of an object.
Early data-driven methods~\cite{kehl2017ssd6d,rad2017bb8,xiang2018posecnn, tekin2018realtime,peng2019pvnet,wang2019densefusion} are trained on task-specific data and show poor generalization to unseen objects, hindering their adoption in real-world applications.
Recent zero-shot approaches address this limitation by leveraging large-scale synthetic training data~\cite{labbe2022megapose,wen2024foundationpose} or foundation models~\cite{ornek2024foundpose,caraffa2024freeze}.
However, the ability of current methods to generalize to realistic industrial settings remains largely unexplored, primarily due to the lack of suitable benchmarks.
Existing datasets often feature objects and environments that differ significantly from those in industrial applications, limiting their relevance.
Most publicly available datasets focus on household items, such as food containers, plastic toys, and office supplies, and are captured in domestic environments~\cite{brachmann2014lmo,kaskman2019hb,xiang2018posecnn}. While a few datasets do include industrial objects, they are typically restricted to plastic electrical components~\cite{hodan2017tless} or metallic mechanical parts~\cite{drost2017itodd,kalra2024ipd}, and are collected in controlled tabletop setups.
This lack of realism is largely due to the challenges of acquiring accurate ground-truth 6D poses, which need controlled capture setups and calibrated turntables~\cite{drost2017itodd,hodan2017tless}.
Such setups offer only limited degrees of freedom and rely on visual markers for calibration, potentially introducing biases in collected data.
To bridge this gap, we pose the question: \textit{Can we collect 6D pose estimation data in realistic industrial environments without relying on such constrained setups?}
Our answer: \textit{Yes, by replacing turntables with robotic arms.}

In this work, we introduce \acronym, a multi-sensor RGBD video dataset of wooden \underline{CH}airs manipulated by a robotic arm in \underline{I}ndustrial scenarios, annotated with precise 6D \underline{P}oses obtained from the robot's kinematics (Fig.~\ref{fig:teaser}, left).
\acronym comprises seven distinct wooden chairs captured using three types of RGBD sensors (LiDAR, passive stereo, and active stereo) and covering different viewpoints (Fig.~\ref{fig:teaser}, right).
The robotic arm's controlled movements provide precise and repeatable object placement, facilitating the generation of datasets with high accuracy and a wide range of poses.
Specifically, \acronym comprises 77,811 RGBD images annotated with ground-truth 6D poses automatically derived from a robot's precisely known joint positions, averaging 11,115 annotations per chair.
\acronym features unique challenges compared to existing 6D pose estimation benchmarks:
(i) a representative industrial environment with realistic clutter, 
(ii) challenging distractor chairs that exhibit subtle differences from the manipulated one, and
(iii) realistic occlusions caused by the robotic arm and human operators interacting with the scene.

We benchmark \acronym using three zero-shot 6D pose estimation methods based on SAM-6D~\cite{lin2024sam6d} and FreeZe~\cite{caraffa2024freeze}.
Our comprehensive analysis evaluates their performance across different sensor types, localization priors, and levels of occlusion.
Although SAM-6D and FreeZe achieve state-of-the-art results on the BOP Benchmark~\cite{bopleaderboard}, their performance on \acronym reveals significant room for improvement, particularly in the presence of noisy depth data, inaccurate localization priors, severe occlusions, and heavy clutter.
This performance gap highlights the complementarity of \acronym to the BOP datasets and reinforces its value as a challenging benchmark for advancing 6D pose estimation in realistic industrial applications.

In summary, our main contributions are:
(i) a multi-sensor 6D pose estimation dataset of wooden chairs, recorded in an industrial setting,
(ii) an automated annotation methodology using a robotic arm for precise ground-truth pose acquisition in realistic environments, and
(iii) a comprehensive evaluation of baseline methods on the proposed dataset, providing a benchmark for future work in this critical area.

%% file: figures/teaser/teaser.tex
\begin{figure}[t]
    \centering
    \begin{overpic}[trim=60 70 0 0, clip, height=40mm]{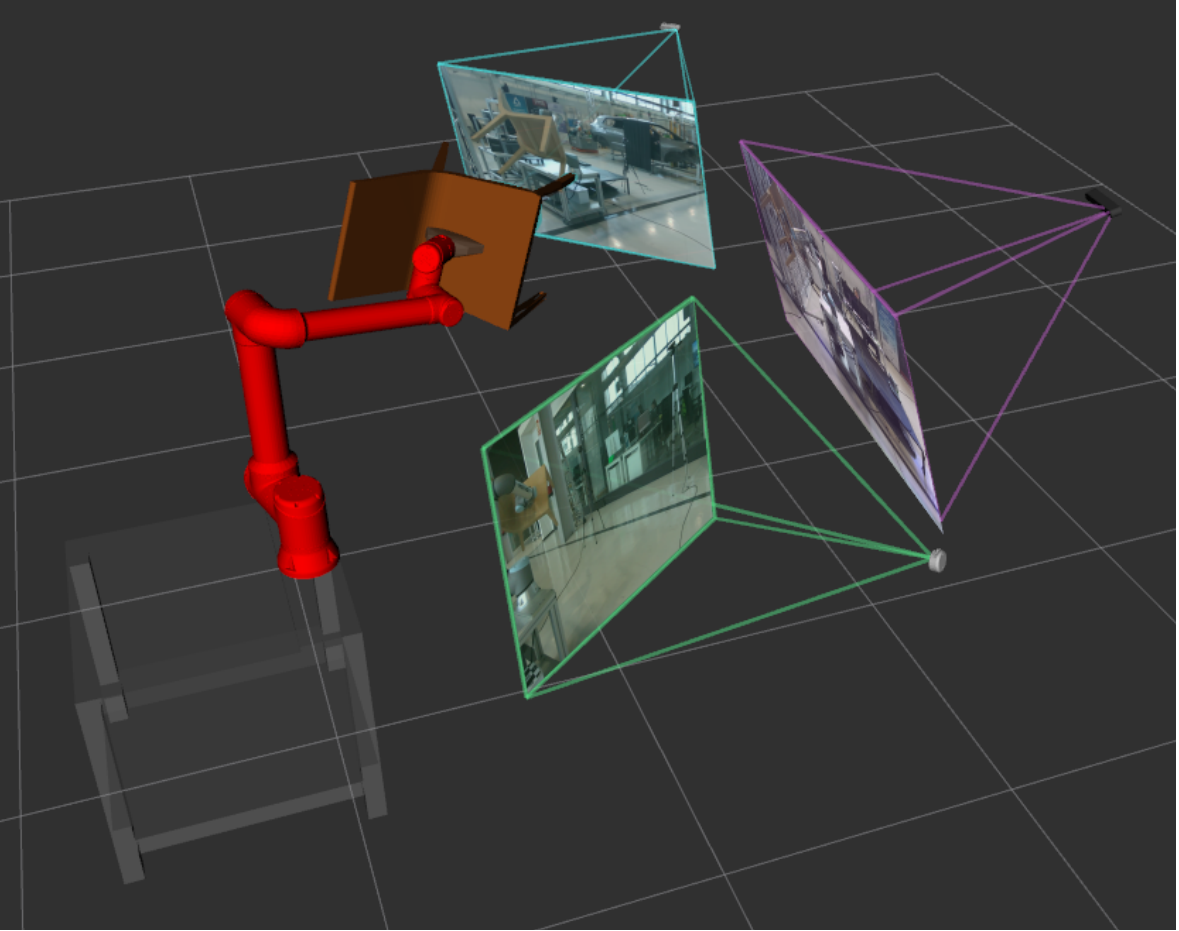}
        \put(15, -5){\footnotesize Acquisition setup (digital twin)}
    \end{overpic}
    \hfill
    \begin{overpic}[height=40mm]{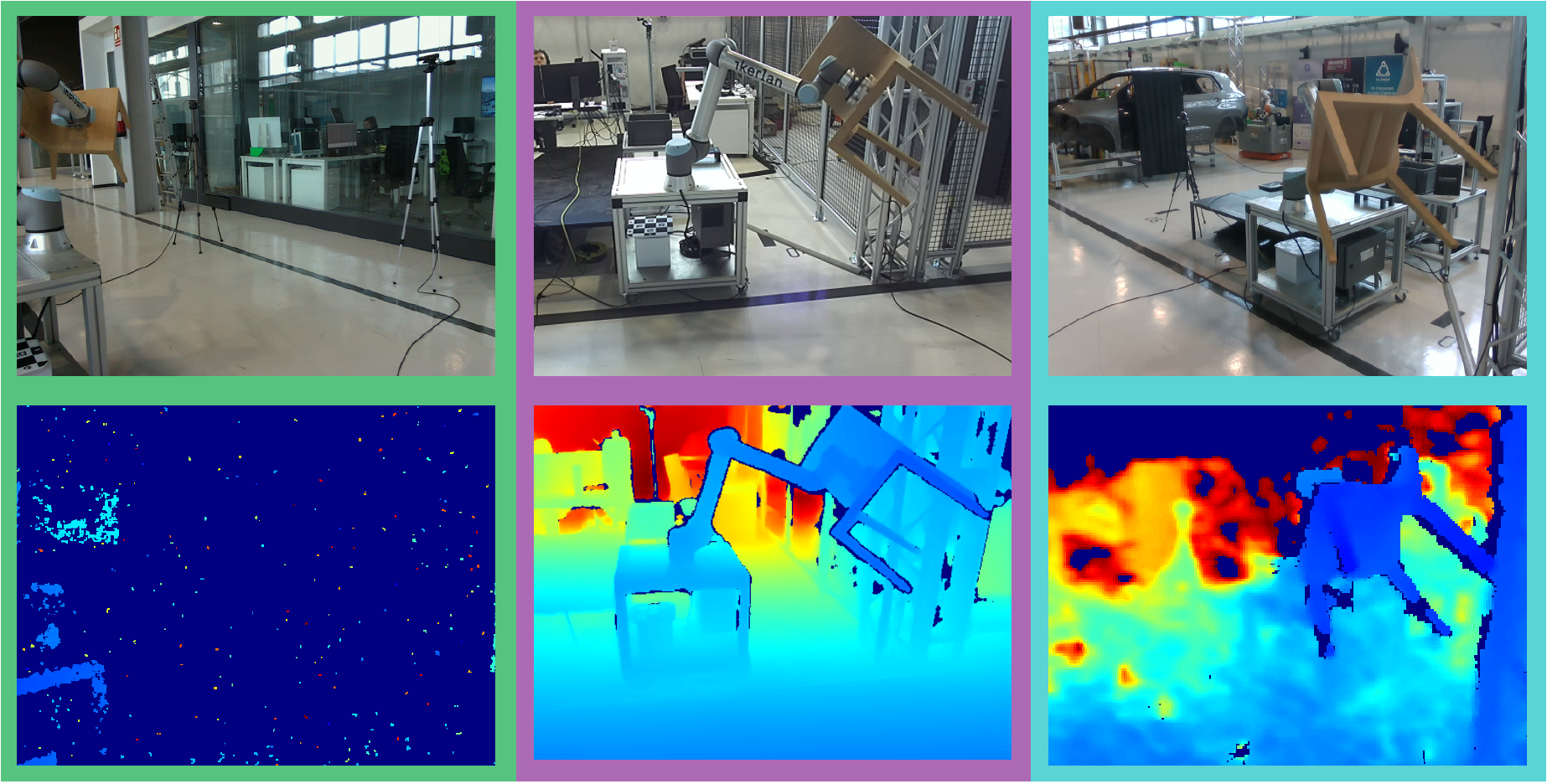}
        \put(-3.5, 8){\footnotesize \rotatebox{90}{Depth}}
        \put(-3.5, 34){\footnotesize \rotatebox{90}{RGB}}
        \put(11, -3.5){\footnotesize LiDAR}
        \put(40, -3.5){\footnotesize Passive stereo}
        \put(73, -3.5){\footnotesize Active stereo}
    \end{overpic}
    
    \caption{
    The \acronym dataset features multi-sensor RGBD videos of a robotic arm manipulating wooden chairs in an industrial setting.
    The images are captured from multiple viewpoints and annotated with ground-truth 6D poses derived from the robot's kinematics, making it a valuable benchmark for evaluating 6D pose estimation methods in realistic scenarios.
    }
    \label{fig:teaser}
\end{figure}
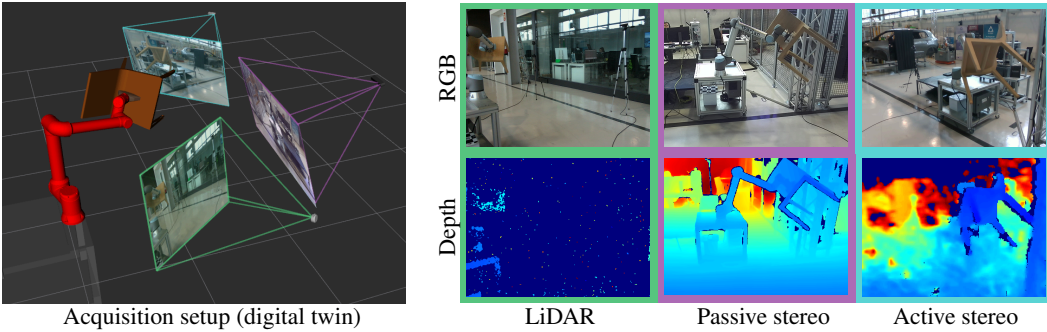

%% file: sections/2_related.tex
\section{Related datasets}\label{sec:related}

\input{tables/related}

6D pose estimation has been studied in both everyday and industrial contexts. 
Everyday objects, used in AR, VR, and human-computer interaction, typically feature rich textures, diverse shapes.
In contrast, industrial objects prioritize functionality over appearance, posing challenges such as texture-less surfaces, reflectivity, high intra-class variation, and stringent precision requirements.
While everyday objects vary greatly in shape and texture, industrial objects often exhibit more uniform geometric structures but differ significantly in materials and finishes.
This work focuses on industrial objects, specifically wooden chairs, which introduce unique challenges due to material and manufacturing similarities.

Existing datasets for 6D pose estimation can be categorized into everyday, industrial, and mixed-object datasets.
Everyday object datasets, such as LM-O~\cite{brachmann2014lmo}, YCB-V~\cite{xiang2018posecnn}, and HOPE-Image~\cite{Tyree2022}, feature diverse household objects with rich textures, often used in cluttered scenes where occlusions and lighting variations impact performance.
Industrial datasets, including T-LESS~\cite{hodan2017tless}, I-TODD~\cite{drost2017itodd}, and MP6D~\cite{chen2022mp6d}, focus on texture-less and metallic objects, emphasizing robustness under challenges like reflectivity and occlusions, often requiring alternative strategies such as shape-based matching.
Hybrid datasets, such as HomebrewedDB\cite{kaskman2019hb}, GraspNet-1Billion\cite{fang2023graspnet}, and TransCG\cite{Fang2022}, contain both everyday and industrial objects, introducing additional variations in lighting, clutter, and transparency, as well as providing grasping data. 
While mixed datasets enable broader generalization, they also introduce inconsistencies in annotation quality and object properties.

Compared to metallic industrial objects, wooden objects pose unique challenges: appearance variability (different wood types, colors, and finishes), texture variations (impacting keypoint-based methods), and structural inconsistencies (manufacturing tolerances and deformations). 
Unlike everyday objects, which may exhibit more consistent textures and controlled variations, wooden objects introduce high intra-class variability, complicating pose estimation.  
Some datasets also integrate furniture and robotic arms but differ in scope and focus~\cite{bhatnagar2022behave,heo2023furniturebench}.

\acronym introduces several key novelties compared to existing 6D pose estimation datasets (see Tab.~\ref{tab:comparison}).
While prior benchmarks focus on everyday items~\cite{hinterstoisser2012lm,brachmann2014lmo,hodan2018tudl,doumanoglou2016icbin,xiang2018posecnn}, plastic electrical components~\cite{hodan2017tless}, or metallic mechanical parts~\cite{drost2017itodd} that are rather small (with diameters ranging from 5 to 20cm) and compact (made by a single part), \acronym targets wooden chairs (Fig.~\ref{fig:chairs}(a)), which have significantly larger dimensions (100-120cm in diameter) and consist of multiple parts (back, seat, legs).
Unlike datasets that rely on objects' textures to guide pose estimation~\cite{doumanoglou2016icbin,kaskman2019hb,xiang2018posecnn}, the chairs in \acronym exhibit unreliable textures: the same CAD model may be manufactured using different wood types, grain patterns, or paint finishes, resulting in inconsistent visual appearances.
Most existing datasets are collected in tabletop scenarios, where objects are randomly arranged to create diverse compositions, induce occlusions, and act as distractors.
In contrast, \acronym is recorded in a real-world industrial setting (Fig.~\ref{fig:chairs}(b)), where a robotic arm manipulates the chairs and a human operator interacts with the scene.
This setup introduces more realistic occlusions and makes the robot a plausible distractor commonly found in industrial automation settings.
Another major distinction is that most datasets derive ground-truth 6D poses using turntables calibrated with visual markers, which may bias the data and introduce artifacts not present in real-world deployment scenarios. 
In contrast, \acronym obtains ground-truth 6D poses by combining robot kinematics with precomputed camera-to-robot calibrations, avoiding any visual markers in the data.
\acronym has a larger working distance (150-600cm), with significant variation in chair-camera distance due to robotic manipulation, contrasting with the limited distance variation in turntable setups.
The dataset most similar to ours is IPD~\cite{kalra2024ipd}, as it also features a robotic arm and industrial objects.
However, IPD is limited to top-down views rather than capturing multiple viewpoints, considers different sensors, and primarily focuses on varying lighting conditions rather than occlusions caused by the robotic arm and human operators.

%% file: tables/related.tex
\newcolumntype{h}{>{\columncolor{myazure}}c}
\begin{table*}[t]
    \centering
    \caption{
    Comparison of object 6D pose estimation datasets in industrial settings: T-LESS, ITODD, IPD, and our proposed dataset, \acronym. Columns represent the different datasets, while rows are organized into three groups—Sensors, Objects, and Scenes—each detailing relevant dataset characteristics such as sensor types, object materials, and scene setups.
    }
    \label{tab:comparison}

    \resizebox{\columnwidth}{!}{%
    \begin{tabular}{llccch}
        \toprule
        & & T-LESS~\cite{hodan2017tless} & ITODD~\cite{drost2017itodd} & IPD~\cite{kalra2024ipd} & \acronym (ours) \\
        \toprule
        \multirow{3}{*}{Sensors} & Types & RGB or D-only & Grayscale or D-only & RGB or RGBD or Polar & RGBD \\
        & Depth & Structured light, TOF & Stereo & Structured light & Active and passive stereo, LiDAR \\
        & Viewpoints & Same & Different & Different & Different \\ 
        \midrule
        \multirow{4}{*}{Objects} & Type & Electrical components & Mechanical parts & Mechanical parts & Chairs \\
        & Material & Plastic & Metallic & Metallic & Wooden \\
        & Texture & Non-informative & Not available & Non-informative & Unreliable \\
        & Size & 5-20cm & 5-20cm & 5-50cm & 100-120cm \\
        \midrule
        \multirow{4}{*}{Scenes} & Setup & Turntable with markers & Turntable with markers & Robotic arm with table & Robotic arm without tables or markers \\
        & Occlusions & Other objects & Other objects & Other objects & Robotic arm, Human operator, Other chair \\
        & Working distance & <100cm & <50cm & 150-200cm & 150-600cm \\
        & Datatype & Images & Images & Images & Images and videos \\
        \bottomrule
    \end{tabular}
    }
\end{table*}

%% file: sections/3_dataset.tex
\section{The \acronym dataset}\label{sec:dataset}

\input{figures/chairs/chairs}

\subsection{Acquisition setup}

\noindent\textbf{Robotic equipment.}
We used a Universal Robots UR10e collaborative arm~\cite{URmanual} because of its long reach and high payload capacity, enabling a safe handling of bulky objects alongside operators.
For reliable grasping, we used an OnRobot VG10 pneumatic vacuum gripper~\cite{OnRobotmanual}, whose adjustable suction strength and dual zones match a wide range of chairs. 
For porous chair surfaces, we glued a small polypropylene pad to the vacuum cup to form a reliable seal.
We implemented a repeatable motion routine for all recordings that sweeps the chair through a wide variety of rotations and angles to capture every facet of the chair.

\noindent\textbf{Vision sensors.}
We used three different RGBD sensors: an Intel \rsd, a StereoLabs \zed, and an Intel \rsl.
The \rsd uses active stereo to capture depth from disparities between two infrared (IR) cameras and an IR projector to enhance depth estimation on low-texture surfaces.
The \zed uses passive stereo to estimate depth from disparities between two RGB cameras.
Both sensors feature global shutters, reducing motion blur from robot movements. 
The \rsl is a solid-state LiDAR sensor that measures depth by emitting laser pulses and capturing reflections with a time-of-flight sensor.
All sensors provide real-time depth, but differ in range: the \rsd operates up to 10m, the \zed up to 20m, and the \rsl achieves millimeter accuracy up to 9m.
Each has an integrated RGB cameras that provide color information to depth-aligned images.
We chose these for Intel RealSense's popularity, the \zed's range, and LiDAR's accuracy.

\noindent\textbf{Calibration.}
Fig.~\ref{fig:chairs}(c) illustrates the calibration pattern used to determinate the transformation from each sensor to the robot base. 
The pattern is mounted on the robot's end-effector at a precisely measured offset. 
The robot positions the pattern in view of the camera, and image processing (e.g., OpenCV) locates its center, yielding the sensor-to-end-effector transform, $\textbf{T}^{S \to E} \in SE(3)$.
The robot's kinematic model provides the end-effector-to-base transform, $\textbf{T}^{E \to B} \in SE(3)$. Combining these two gives the sensor-to-base transformation, $\textbf{T}^{S \to B} \in SE(3)$.
Next, the robot computes the base-to-chair transform, $\textbf{T}^{B \to C} \in SE(3)$, using its kinematics to know the end-effector pose relative to the base, and then a tool-center-point (TCP) calibration is performed with the chair attached as a ``tool''.
Using a four-point method—approaching the same feature on the chair (for example, an edge of the backrest) from four distinct orientations the exact end-effector-to-chair transform, $\textbf{T}^{E \to C} \in SE(3)$, is found. 
Finally, by chaining the sensor-to-base, base-to-end-effector, and end-effector-to-chair transforms, $\textbf{T}^{S \to B}$, $\textbf{T}^{B \to E}$ and $\textbf{T}^{E \to C}$, the system obtains the overall sensor-to-chair transformation, $\textbf{T}^{S \to C} \in SE(3)$, ensuring that all depth and color data are accurately referenced to the chair's coordinate frame.
Fig.~\ref{fig:chairs}(d) illustrates the obtained transformations, color-coded as: sensor-to-base (green), base-to-end-effector (azure), end-effector-to-chair (yellow), and sensor-to-chair (white).

\subsection{Data}

\noindent\textbf{RGBD images.}
The \acronym dataset provides 77,811 RGBD frames with corresponding annotations: 19,537 from the \rsd, 39,547 from the \rsl, and 18,727 from the \zed.
It includes RGBD images captured from three different viewpoints.
Each frame includes the 6D pose of the chair relative to the camera, computed by combining the robot-to-camera transformation from calibration and the chair-to-robot transformation from the robot’s kinematics.
Our approach facilitates a comparative evaluation of image quality and depth reliability across different sensing technologies.

\noindent\textbf{3D chair models.}
The \acronym dataset features seven chair models: three solid-wood and four frame-only designs, originally including cushions. The chairs are bonded with adhesives and screws, and their finishes range from natural wood to painted.
While the original CAD files (available from Andreu World catalog~\cite{andreuworld} in ``.dwg'' format for AutoCAD 2007+) include both rigid wooden parts and soft elements, we have removed all non-wooden components using Blender~\cite{blender} to produce clean, texture-less models focused solely on structural geometry.

\noindent\textbf{Core features.}
\noindent Fig.~\ref{fig:pose_errors} compares the 6D pose distributions between the T-LESS~\cite{hodan2017tless} dataset and \acronym.
It shows the distribution of residuals for rotation (left) and translation (right) components, where residuals are computed as deviations from the mean pose.
For T-LESS, we report results for object ID 9 in scene 12, while for \acronym we use a representative full chair model. In both cases, the plots aggregate data from three sensors, using 50 frames per camera.
The results indicate that both datasets show similar distributions for pitch and yaw angles; however, \acronym exhibits greater variability in the roll angle.
\acronym exhibits significantly higher variability in translation: while T-LESS objects typically move within a 100\,mm range, the objects in \acronym can undergo translations of up to 2000\,mm along each axis.
This highlights the more dynamic and challenging nature of the \acronym dataset in terms of object movements and spatial variations.
The robotic arm's ability to perform more extensive rotations and translations, beyond turntable limitations, is the primary reason for this enhanced variability.

\input{figures/Data_statistics/translation_rotation}

%% file: figures/chairs/chairs.tex
\begin{figure}[t]
    \centering
    
    \begin{overpic}[trim=90 0 50 200, clip, height=26mm]{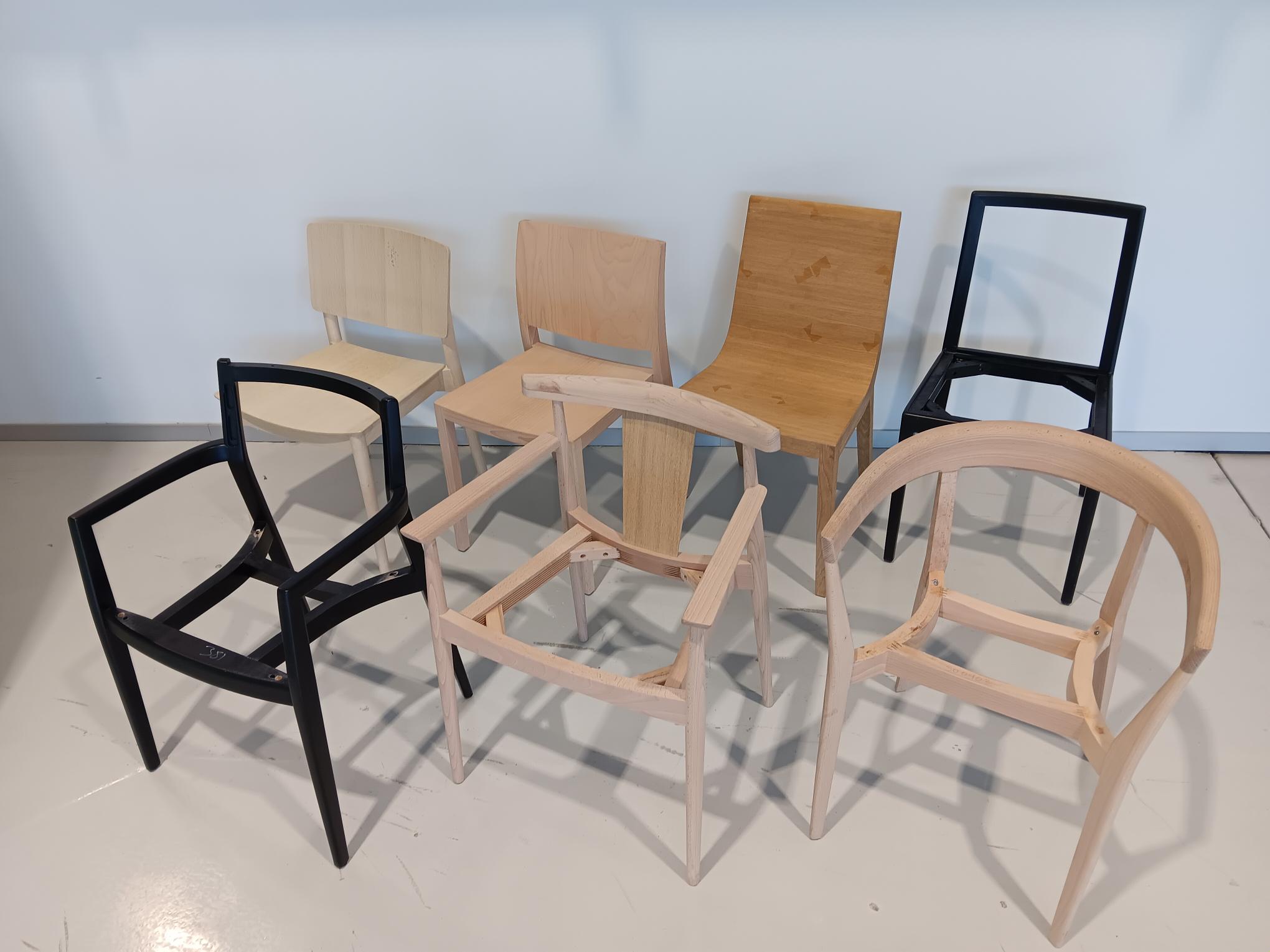}
    \put(-2.5, 3){\colorbox{white}{(a)}}
    \end{overpic}
    \begin{overpic}[trim=80 0 130 0, clip, height=26mm]
    {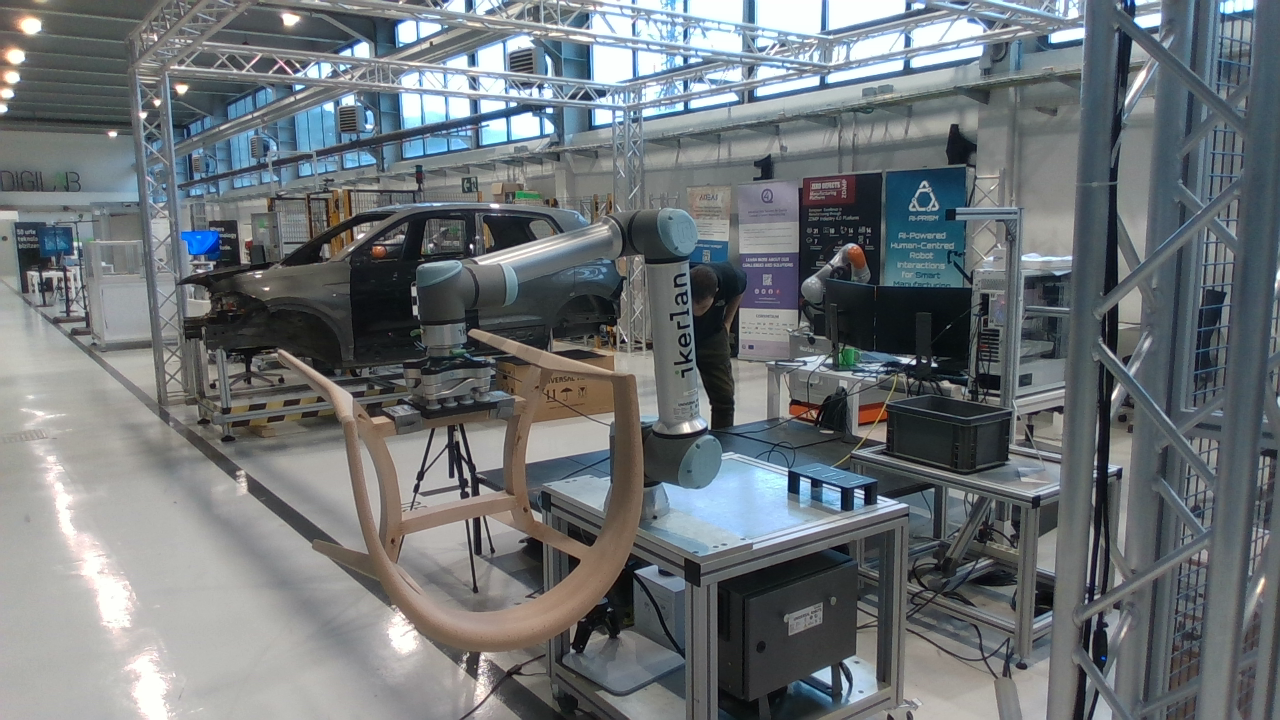}
    \put(-2.5, 3){\colorbox{white}{(b)}}
    \end{overpic}
    \begin{overpic}[trim=0 60 30 60, clip, height=26mm]{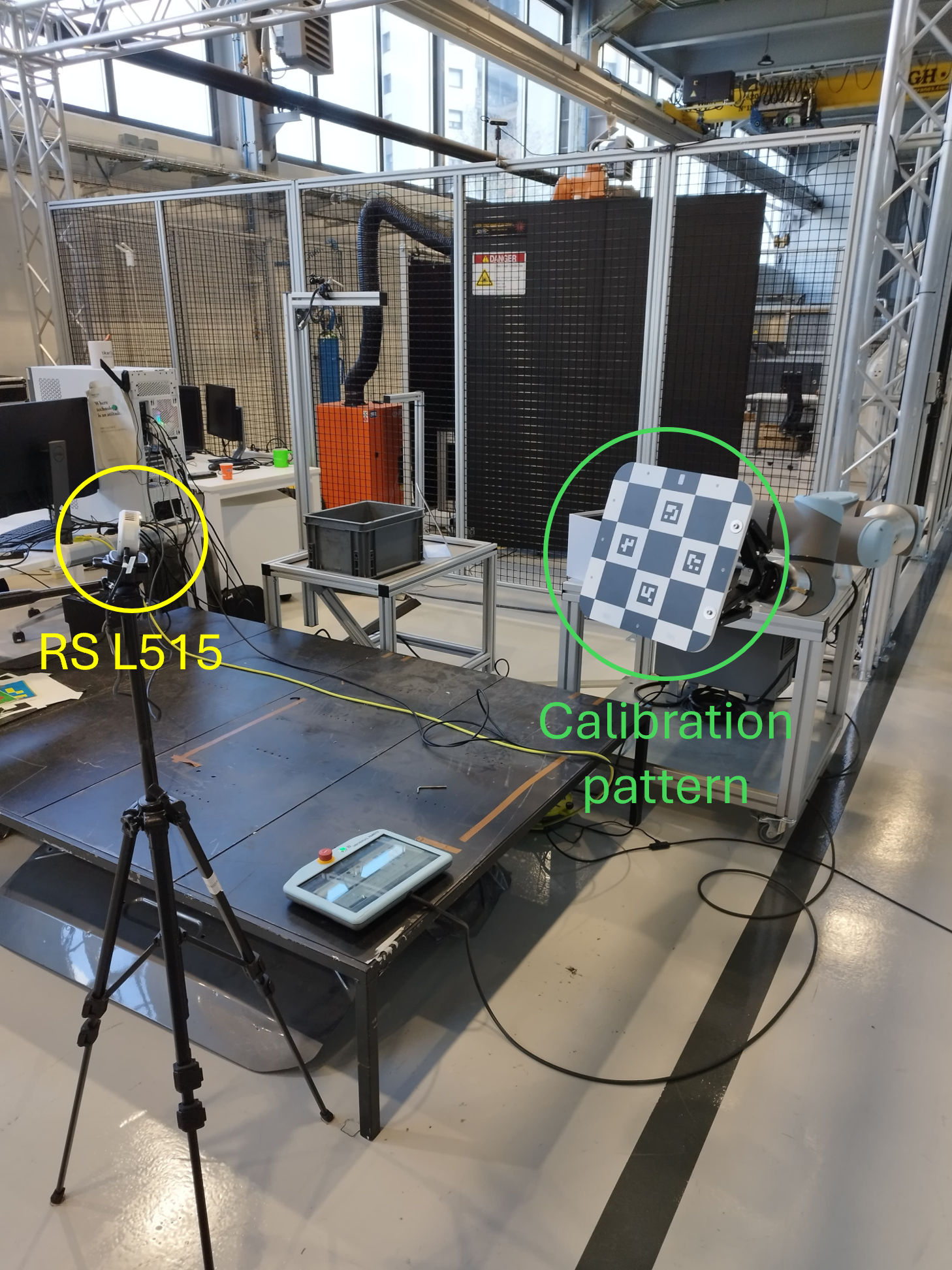}
    \put(-2.5, 3){\colorbox{white}{(c)}}
    \end{overpic}
    \begin{overpic}[trim=0 100 250 0, clip, height=26mm]{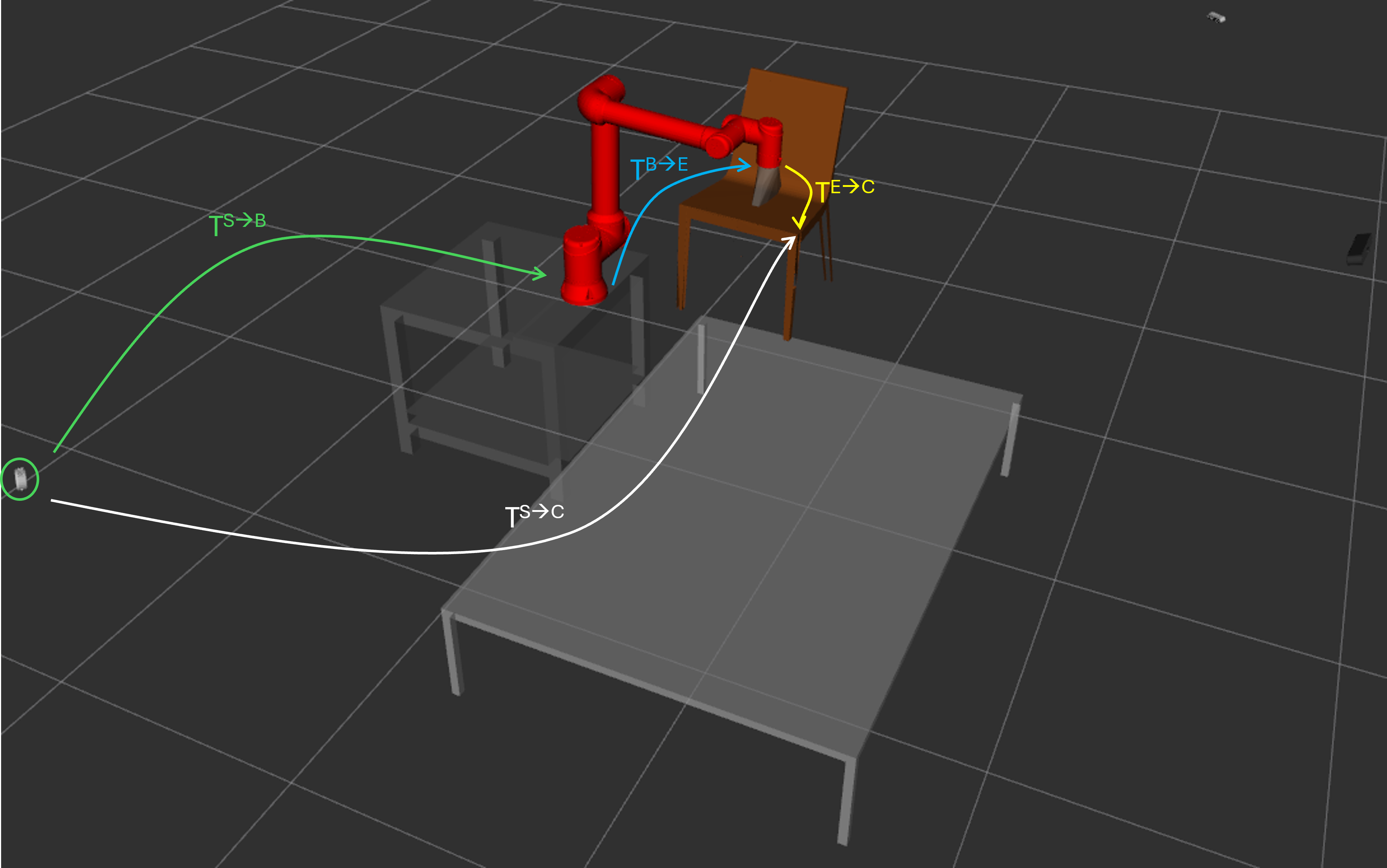}
    \put(-2.5, 3){\colorbox{white}{(d)}}
    \end{overpic}

    \vspace{-3mm}
    \caption{
    (a) Manufactured wooden chairs included in \acronym.
    (b) Industrial environment used to record \acronym, featuring the robotic arm, the end-effector, and realistic clutter.
    (c) \rsl calibration setup.
    (d) Transformations involved during the calibration procedure.
    }
    \label{fig:chairs}
\end{figure}

%% file: figures/Data_statistics/translation_rotation.tex
\begin{figure}[t]
    \centering

    \begin{minipage}[b]{0.48\textwidth}
        \centering
        \footnotesize Rotation
    \end{minipage}
    \hfill
    \begin{minipage}[b]{0.48\textwidth}
        \centering
        \footnotesize Translation
    \end{minipage}
    \vspace{-1mm}

    \begin{minipage}[b]{0.24\textwidth}
        \centering
        \footnotesize T-LESS~\cite{hodan2017tless}
    \end{minipage}
    \begin{minipage}[b]{0.24\textwidth}
        \centering
        \footnotesize \acronym (ours)
    \end{minipage}
    \hfill
    \begin{minipage}[b]{0.24\textwidth}
        \centering
        \footnotesize T-LESS~\cite{hodan2017tless}
    \end{minipage}
    \begin{minipage}[b]{0.24\textwidth}
        \centering
        \footnotesize \acronym (ours)
    \end{minipage}

    \begin{minipage}[b]{0.24\textwidth}
        \centering
        \begin{overpic}[trim=0 0 10 0, clip, height=17mm]{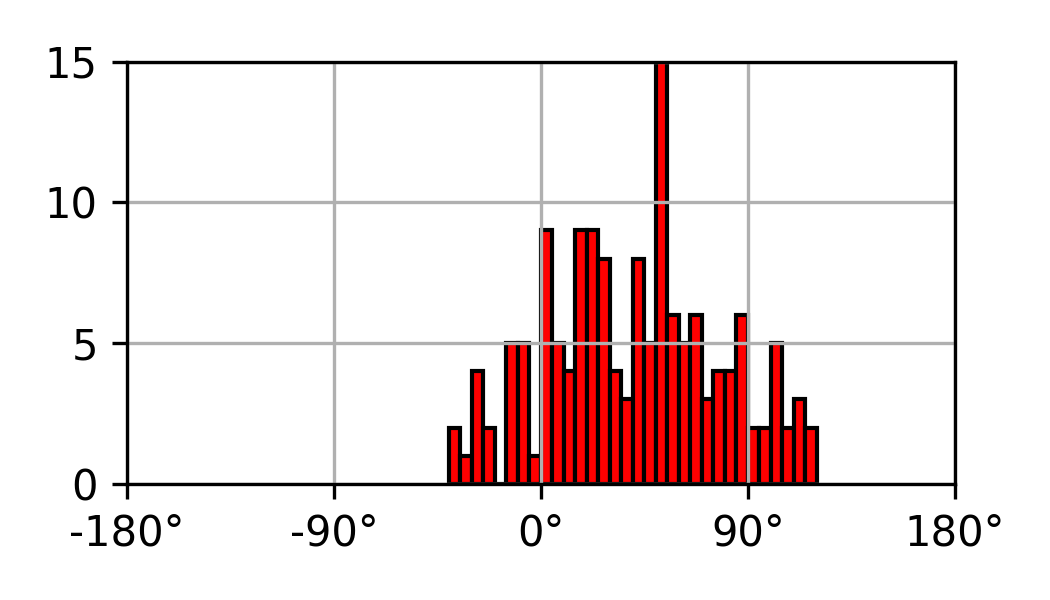}
            \put(-5,25){\footnotesize \rotatebox{90}{Roll}}
        \end{overpic}
    \end{minipage}
    \begin{minipage}[b]{0.24\textwidth}
        \centering
        \includegraphics[trim=5 0 5 0, clip, height=17mm]{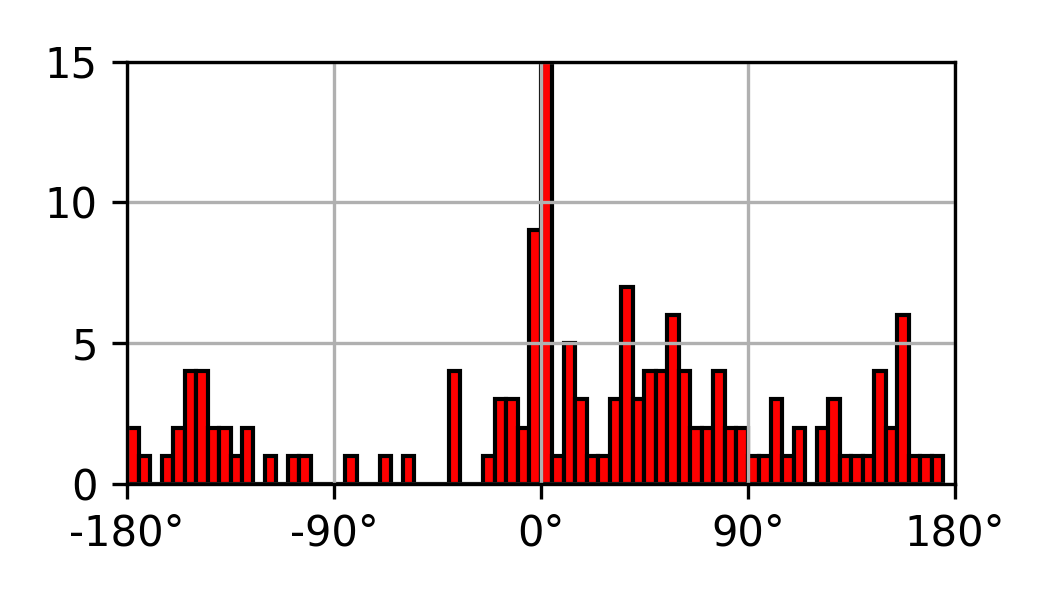}
    \end{minipage}
    \hfill
    \begin{minipage}[b]{0.24\textwidth}
        \centering
        \begin{overpic}[trim=0 0 10 0, clip, height=17mm]{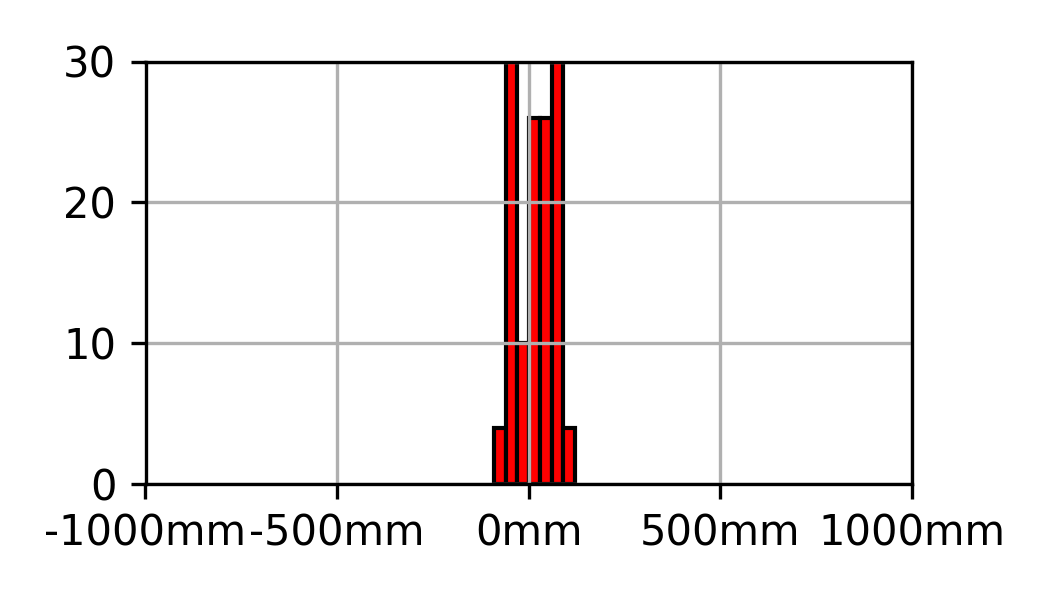}
            \put(-5,30){\rotatebox{90}{\footnotesize X}}
        \end{overpic}
    \end{minipage}
    \begin{minipage}[b]{0.24\textwidth}
        \centering
        \includegraphics[trim=5 0 5 0, clip, height=17mm]{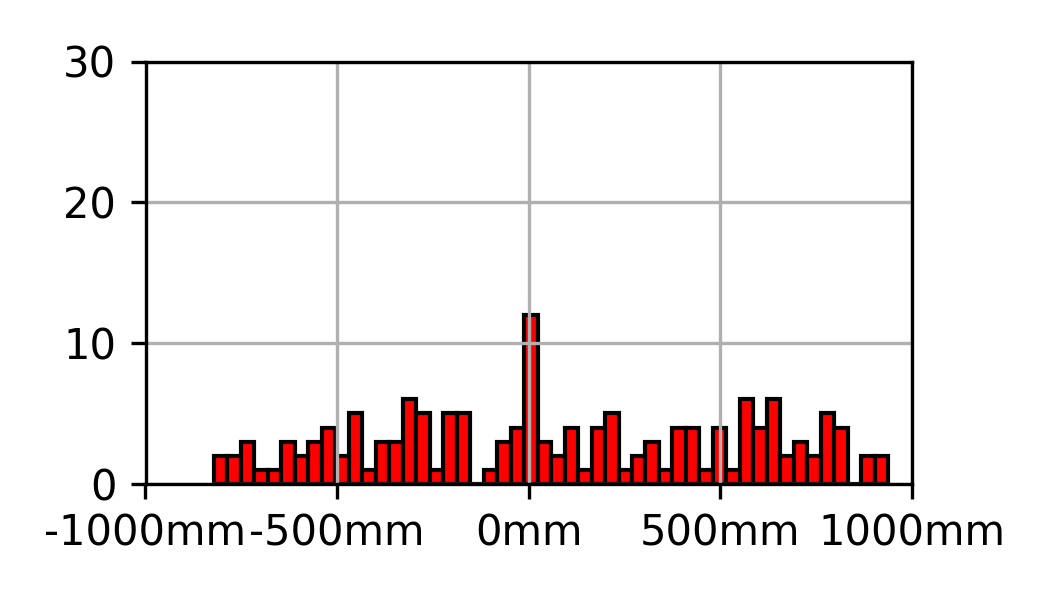}
    \end{minipage}

    \vspace{-3mm}

    \begin{minipage}[b]{0.24\textwidth}
        \centering
        \begin{overpic}[trim=0 0 10 0, clip, height=17mm]{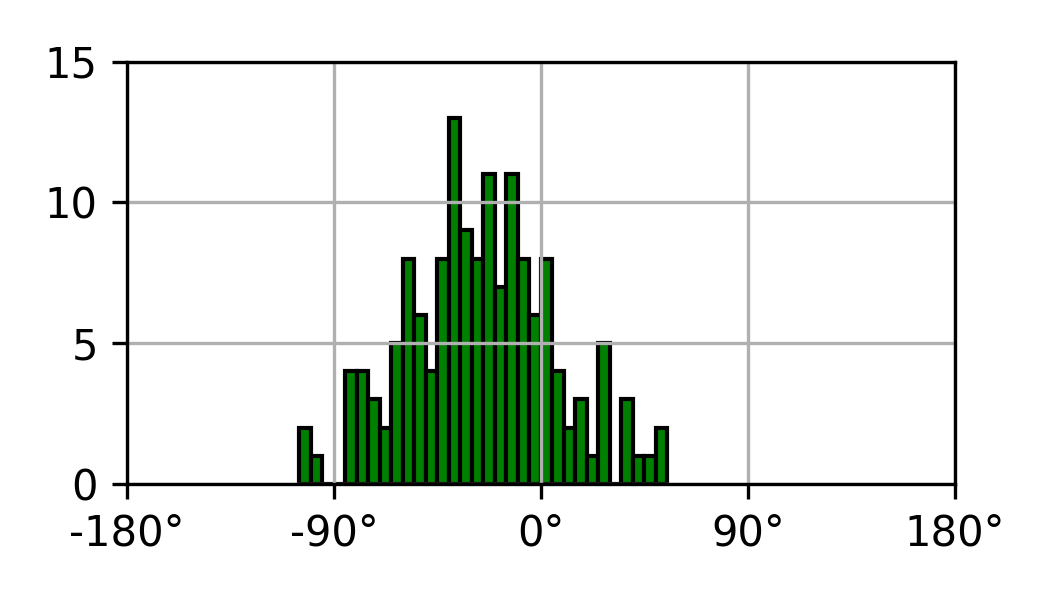}
            \put(-5,25){\rotatebox{90}{\footnotesize Pitch}}
        \end{overpic}
    \end{minipage}
    \begin{minipage}[b]{0.24\textwidth}
        \centering
        \includegraphics[trim=5 0 5 0, clip, height=17mm]{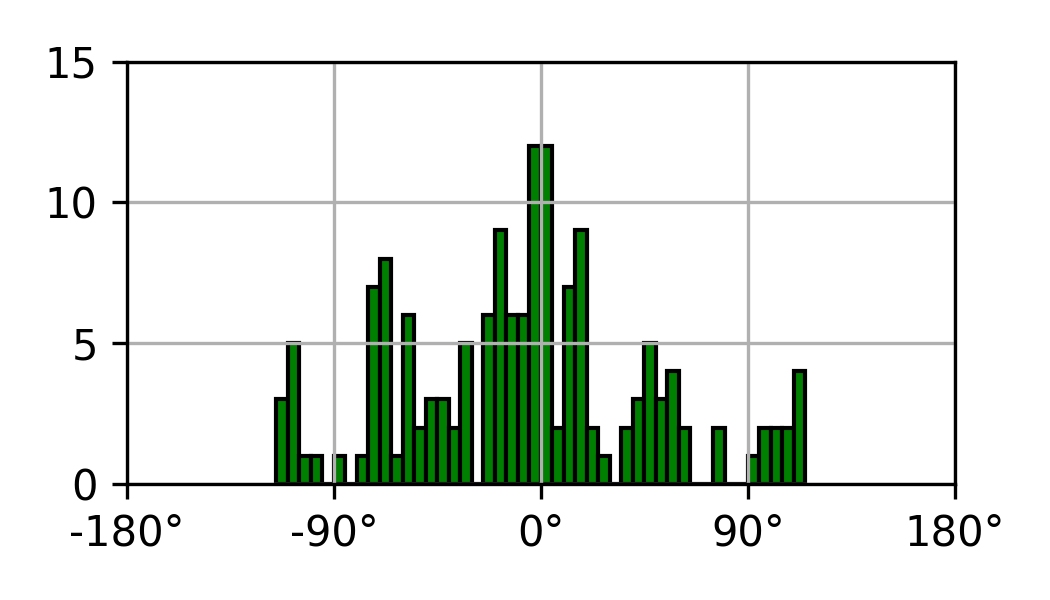}
    \end{minipage}
    \hfill
    \begin{minipage}[b]{0.24\textwidth}
        \centering
        \begin{overpic}[trim=0 0 10 0, clip, height=17mm]{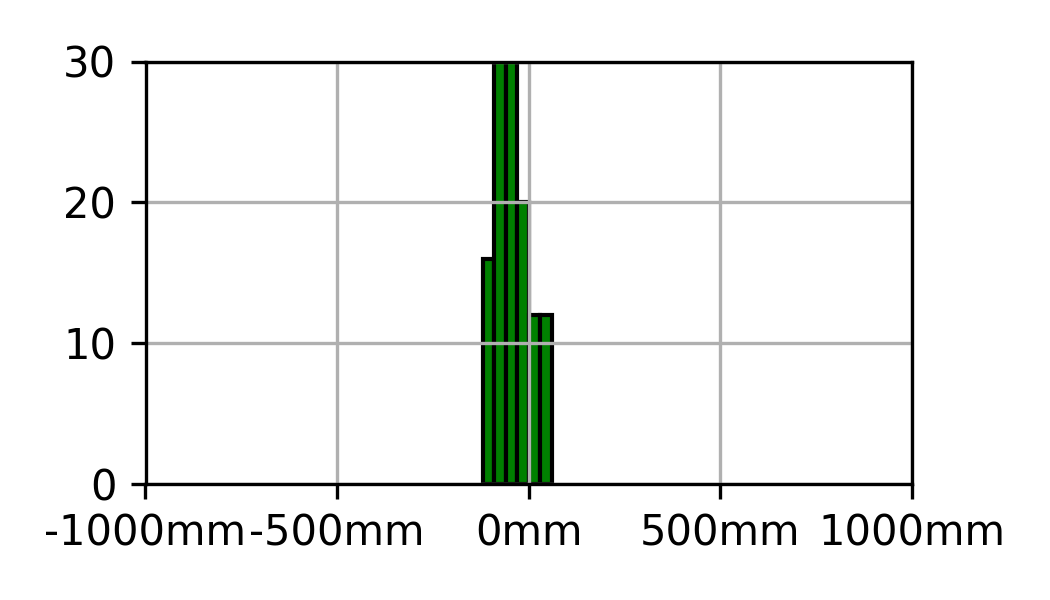}
            \put(-5,30){\rotatebox{90}{\footnotesize Y}}
        \end{overpic}
    \end{minipage}
    \begin{minipage}[b]{0.24\textwidth}
        \centering
        \includegraphics[trim=5 0 5 0, clip, height=17mm]{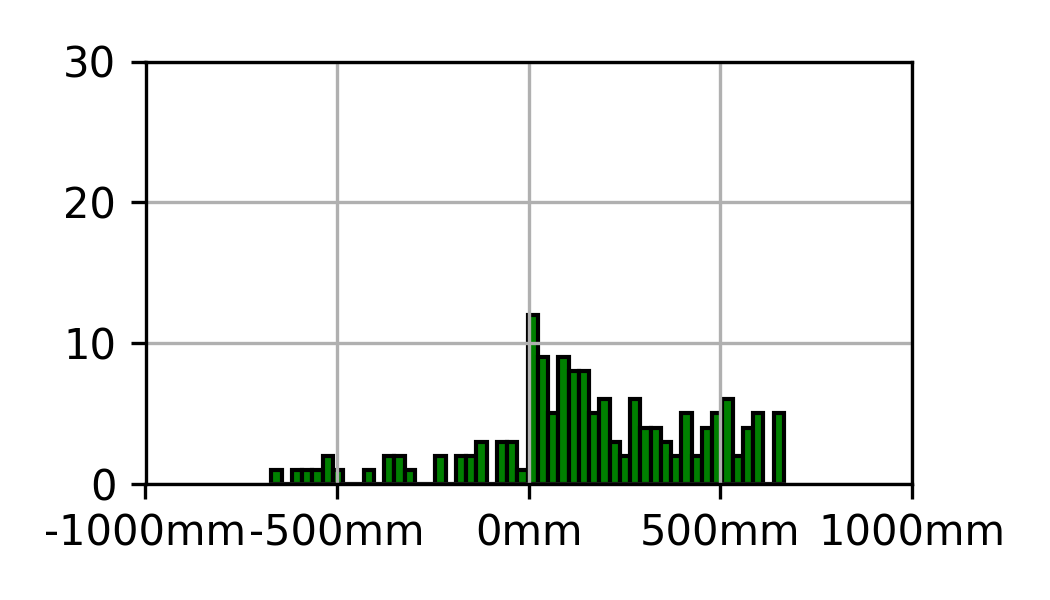}
    \end{minipage}

    \vspace{-3mm}

    \begin{minipage}[b]{0.24\textwidth}
        \centering
        \begin{overpic}[trim=0 0 10 0, clip, height=17mm]{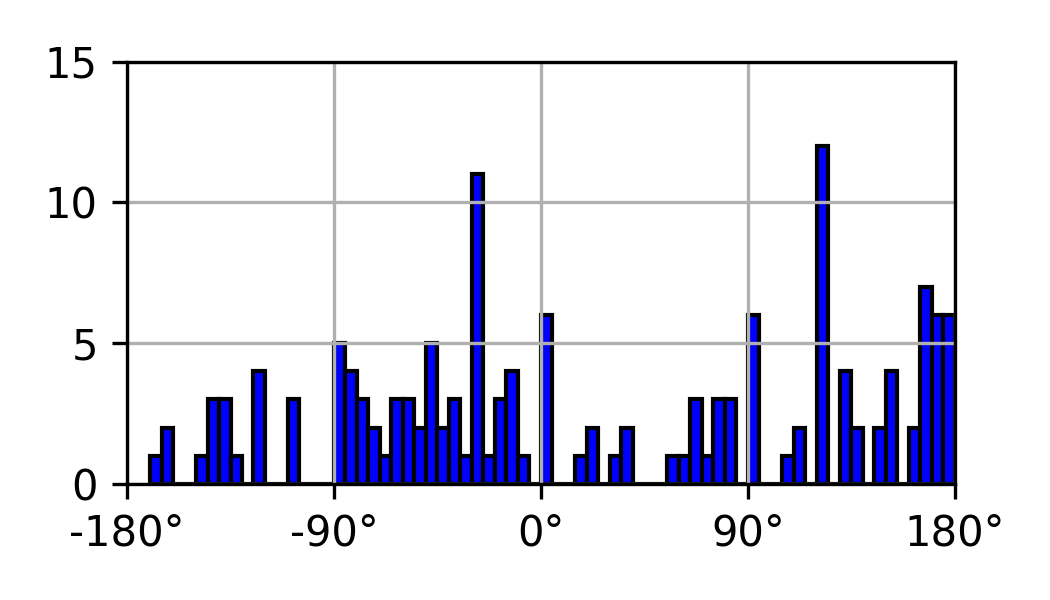}
            \put(-5,25){\footnotesize \rotatebox{90}{Yaw}}
        \end{overpic}
    \end{minipage}
    \begin{minipage}[b]{0.24\textwidth}
        \centering
        \includegraphics[trim=5 0 5 0, clip, height=17mm]{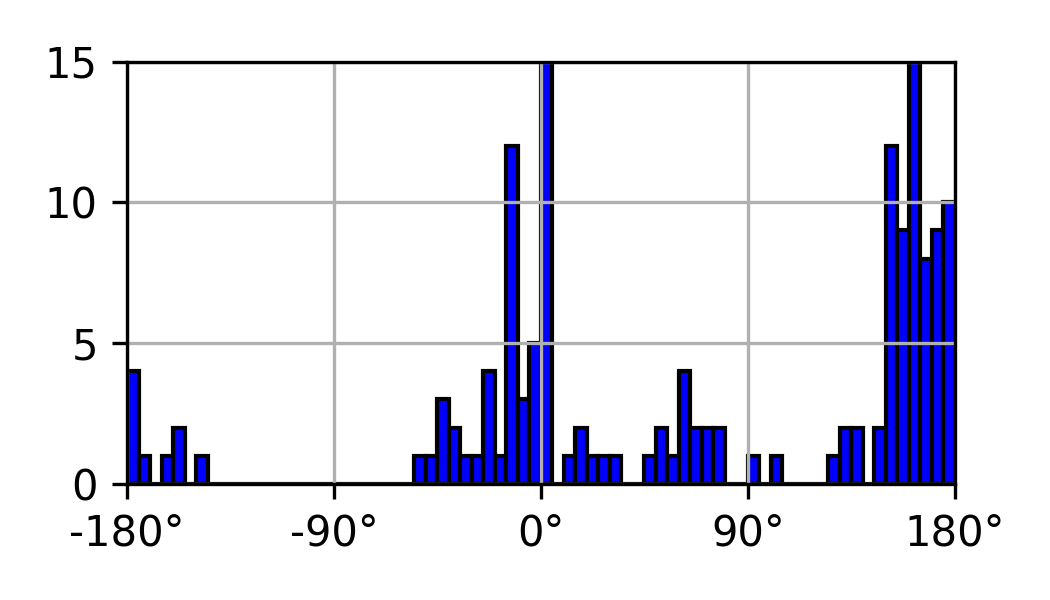}
    \end{minipage}
    \hfill
    \begin{minipage}[b]{0.24\textwidth}
        \centering
        \begin{overpic}[trim=0 0 10 0, clip, height=17mm]{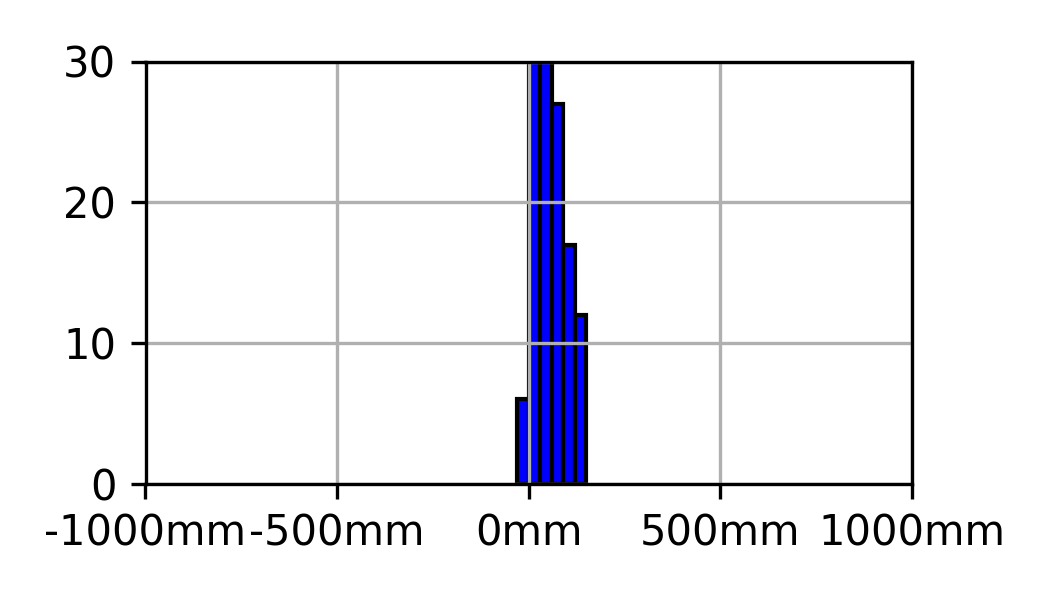}
            \put(-5,30){\footnotesize \rotatebox{90}{Z}}
        \end{overpic}
    \end{minipage}
    \begin{minipage}[b]{0.24\textwidth}
        \centering
        \includegraphics[trim=5 0 5 0, clip, height=17mm]{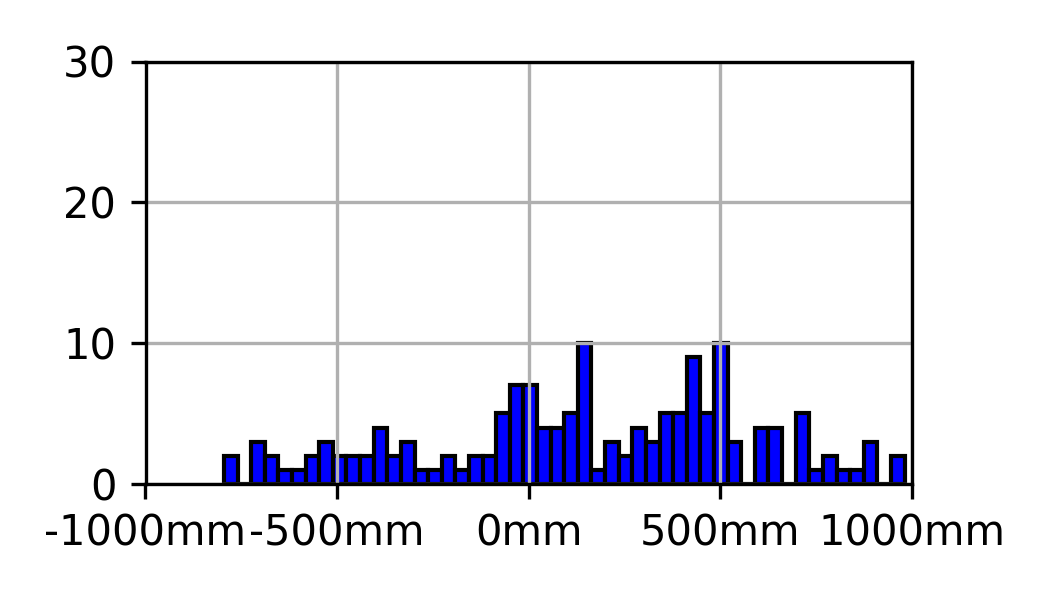}
    \end{minipage}

    \vspace{-4mm}
    \caption{
    Comparison between T-LESS~\cite{hodan2017tless} and \acronym ground-truth 6D poses.
    Histograms show the distribution of residuals for rotation (left) and translation (right) components. 
    \acronym shows significantly higher variability in terms of roll angles (top left) and translations (right).
    }
    \label{fig:pose_errors}
\end{figure}

%% file: sections/4_results.tex
\section{Experimental results}\label{sec:results}

\subsection{6D pose estimation methods}

We benchmark \acronym using three zero-shot 6D pose estimation methods.
The first baseline is SAM-6D~\cite{lin2024sam6d}, which generalizes to unseen objects via large-scale training on task-specific synthetic data~\cite{labbe2022megapose,chang2015shapenet,downs2022google}.
We use the model pretrained on MegaPose~\cite{labbe2022megapose}, as provided in the official implementation.
The other two baselines are FreeZe-GeDi and FreeZe-FPFH.
They are based on FreeZe~\cite{caraffa2024freeze}, a training-free method that integrates geometric and vision foundation models with a RANSAC-based registration pipeline.
FreeZe achieves state-of-the-art performance on the BOP Benchmark~\cite{nguyen2025bop} without requiring task-specific finetuning.
As the original implementation is not publicly available, we reimplemented it.
In our setup, we discard the visual encoder and use either a frozen GeDi~\cite{poiesi2023gedi} pretrained on 3DMatch~\cite{zeng20173dmatch} or the handcrafted FPFH~\cite{rusu2009fpfh} descriptor as the geometric encoder.
For each method, we evaluate two variants: with and without localization priors.
When no localization prior is used, pose estimation is performed on the full RGBD image.
Otherwise, we apply the zero-shot segmentation module of SAM-6D~\cite{lin2024sam6d,kirillov2023sam} to segment the manipulated chair, retain only the most confident mask, and restrict pose estimation to the corresponding segmented region.
We also consider the ground truth chair segmentation as an oracle localization prior to estimate the upper bound on performance in the case of ideal localization.

\subsection{Quantitative results}

We analyze the results obtained with the three baselines on \acronym, reporting rotation and translation errors as defined in ITODD~\cite{drost2017itodd}.
Let $\textbf{T} = (\textbf{R} \in SO(3), \textbf{t} \in \mathbb{R}^3)$ denote a 6D pose, with rotation and translation components $\textbf{R}, \textbf{t}$.
The rotation error measures the angle between the predicted and ground-truth rotation matrices $\textbf{R}_\text{pred}, \textbf{R}_\text{gt}$, and is defined as $E^R = \operatorname*{arccos} \left( ( \operatorname*{Tr} ( \textbf{R}_\text{pred}\textbf{R}_\text{gt}^{-1}) - 1 ) / 2 ) \right)$, where $\operatorname*{Tr}$ denotes the trace operator.
The translation error measures the Euclidean distance between the object barycenter transformed by the predicted and ground-truth 6D poses $\textbf{T}_\text{pred}, \textbf{T}_\text{gt}$, and is defined as $E^T = \lVert \textbf{T}_\text{pred} \textbf{b} - \textbf{T}_\text{gt} \textbf{b} \rVert_2$, where $\textbf{b} = (1/N \sum_{n=1}^N x_n, 1/N \sum_{n=1}^N y_n, 1/N \sum_{n=1}^N z_n, 1)$ is the object barycenter expressed in homogeneous coordinates, and $(x_n, y_n, z_n) \in \mathbb{R}^3$ are the 3D coordinates of the object points.
$E^R$ is measured in degrees, while $E^T$ is measured in millimeters.
For each error type, we first compute the average across all images containing a given chair type and then report the mean and standard deviation across all chair types.

\input{tables/quant_noocc}
\input{tables/quant_occ}

Tables~\ref{tab:quant_noocc} and \ref{tab:quant_occ} compare $E^R$ and $E^T$ across different sensors (columns) and localization priors (rows): no prior, zero-shot segmentation (ZS), and ground-truth segmentation (GT).
Tab.~\ref{tab:quant_noocc} focuses on the portion of \acronym that is not affected by human-induced occlusions or the presence of distractor chairs, while Tab.~\ref{tab:quant_occ} reports results on the most difficult subset of data, featuring both of these challenges.

In general, the FreeZe-GeDi method outperforms the other baselines, demonstrating more accurate pose estimations and better robustness to occlusions and distractors.
FreeZe-FPFH achieves lower performance using the same pipeline, revealing the important role of discriminative geometric features in accurate pose estimation.
SAM-6D produces reasonably accurate rotations, but struggles with translations, where it yields the worst performance overall.
We attribute this to a domain gap, as the chairs in \acronym are out-of-distribution with respect to the synthetic data used to train SAM-6D.
This interpretation is further supported by the fact that, compared to full chairs, which are more common in synthetic datasets, SAM-6D performs 246\% worse on frame-only chairs, whose structure is less conventional.

Pose predictions computed over the entire image are less accurate than those estimated from only the region identified by the localization priors.
This happens because, when the segmentation mask is inaccurate or wider, pose estimation is challenged by partiality or clutter, respectively.
When comparing localization priors, zero-shot segmentation masks underperform ground truth masks, suggesting that the quality of the segmentation masks is crucial for reliable pose estimation.
The SAM-6D segmentor used in our experiments struggles with \acronym chairs, especially for the images recorded with the \rsd sensor, which contains more background clutter and whose depth estimates are noisier compared to the other sensors.
Instead, the \rsl sensor is less affected by localization priors, as its depth information is concentrated on foreground objects, such as the chair and robotic arm.

Tab.~\ref{tab:quant_occ} reports higher errors compared to Tab.~\ref{tab:quant_noocc}, indicating that the baselines are significantly challenged by human-induced occlusions and the presence of distractor chairs.
This happens because the zero-shot segmentor struggles with the partial visibility of the manipulated chair and often fails to identify the correct chair when multiple ones are present.

\subsection{Qualitative results}

\input{figures/qualitative_results/qualitative_results}

Fig.~\ref{fig:qualitative_results} presents qualitative results for FreeZe-GeDi on solid chairs (left half) and frame-only chairs (right half).
We consider three challenges: no occlusions (top), presence of a distractor chair (middle), and mild human-induced occlusion (bottom).
For each setting, we predict the 6D pose of the chair from the entire image (leftmost column) or from the region identified by the SAM-6D zero-shot segmentation (rightmost column, the segmentation mask is shown in the bottom-left corner).
For each image, we overlay the chair CAD model transformed according to the predicted 6D pose, and report $E^R$ and $E^T$ to facilitate comparison.

The top part shows images captured with the \rsd sensor, without severe occlusions.
In the first row, the localization prior fails to identify the chair and instead segments background clutter, resulting in an incorrect 6D pose estimation.
The second row highlights two notable edge cases that illustrate how accurate localization does not always ensure successful pose recovery.
In one case, the predicted pose is inaccurate despite a precise segmentation; in the other, the pose is accurately estimated despite poor segmentation.
The middle part shows images captured with the \rsl sensor, featuring an additional chair acting as a distractor.
When the localization prior segments the distractor chair instead of the manipulated one, accurate pose estimation becomes infeasible.
In the fourth row, a precise segmentation significantly improves pose estimation accuracy compared to the case without any prior.
The segmentation of the frame-only chair is less affected by the presence of the distractor chair, likely due to the more pronounced differences between them.
The bottom part shows images captured with the \zed sensor, affected by human-induced occlusions.
When the segmentation of the partially occluded chair is accurate, the resulting pose estimation also improves and outperforms the prediction obtained without any localization prior. 
In general, frame-only chairs yield higher errors, as their thin structures lead to less reliable segmentation masks and noisier depth measurements, both of which negatively impact the subsequent pose estimation.

These examples illustrate that, although the FreeZe-GeDi method can perform well in certain cases, its accuracy remains highly sensitive to the quality of localization, the presence of occlusions and distractors, and the specific characteristics of the object.

%% file: tables/quant_noocc.tex
\newcolumntype{h}{>{\columncolor{myazure}}c}
\begin{table*}[t!]
\centering

\caption{
Quantitative results for non-occluded videos, reporting rotation and translation errors while comparing different sensors and localization priors.
}
\label{tab:quant_noocc}

\resizebox{\columnwidth}{!}{%
\begin{tabular}{rlchhcchh}
    \toprule
    & \multirow{2}{*}{Method} & Localization & \multicolumn{2}{c}{\cellcolor{myazure}\zed} & \multicolumn{2}{c}{\rsd} & \multicolumn{2}{c}{\cellcolor{myazure}\rsl} \\
    & & prior & $E^R$ & $E^T$ & $E^R$ & $E^T$ & $E^R$ & $E^T$ \\
    \toprule
    {\color{gray} \scriptsize 1} & \multirow{1}{*}{FreeZe-FPFH~\cite{rusu2009fpfh}} & & $129 \pm 36$ & $1807 \pm 454$ & $126 \pm 36$ & $1489 \pm 580$ & $134 \pm 45$ & $516 \pm 393$ \\
    {\color{gray} \scriptsize 2} & \multirow{1}{*}{SAM-6D~\cite{lin2024sam6d}} & & $119 \pm 38$  &$2099 \pm 240$ & $127 \pm 36$ & $5130 \pm 728$ & $125 \pm 38$ & $ 1433 \pm 366 $        \\
    {\color{gray} \scriptsize 3} & \multirow{1}{*}{FreeZe-GeDi~\cite{poiesi2023gedi}} & & $83 \pm 64$ & $552 \pm 674$ & $103 \pm 58$ & $1002 \pm 810$ & $46 \pm 44$ & $185 \pm 183$ \\
    \midrule
    {\color{gray} \scriptsize 4} & \multirow{1}{*}{FreeZe-FPFH~\cite{rusu2009fpfh}} & ZS  & $ 122\pm42 $ & $ 1402\pm 658$ & $124 \pm 41$ & $1158 \pm 635 $ & $ 116\pm 50 $ & $504 \pm 366$ \\
    {\color{gray} \scriptsize 5} & \multirow{1}{*}{SAM-6D~\cite{lin2024sam6d}} & ZS & $ 92\pm 49$ & $ 2066\pm 1441$ & $ 90\pm 45$ & $5427 \pm 4754$ & $ 75\pm 45 $ & $886 \pm 750$ \\
    {\color{gray} \scriptsize 6} & \multirow{1}{*}{FreeZe-GeDi~\cite{poiesi2023gedi}} & ZS & $ 75\pm 58$ & $494 \pm 750$ & $ 97\pm 55$ & $ 693\pm 762$ & $ 50\pm 50$ & $204 \pm 263$ \\
    \midrule
    {\color{gray} \scriptsize 7} & \multirow{1}{*}{FreeZe-FPFH~\cite{rusu2009fpfh}} & GT & $113 \pm 55$ & $155 \pm 80$ & $113 \pm 54$ & $193 \pm 95$ & $114 \pm 55$ & $213 \pm 181$ \\
    {\color{gray} \scriptsize 8} & \multirow{1}{*}{SAM-6D~\cite{lin2024sam6d}} & GT & $27 \pm 41$ & $114 \pm 171$ & $80 \pm 55$ & $1260 \pm 1164$ & $52 \pm 47$ & $478 \pm 410$\\
    {\color{gray} \scriptsize 9} & \multirow{1}{*}{FreeZe-GeDi~\cite{poiesi2023gedi}} & GT & $27 \pm 46$ & $54 \pm 46$ & $61 \pm 68$ & $107 \pm 72$ & $38 \pm 38$ & $100 \pm 112$ \\
    \bottomrule
\end{tabular}
}
\end{table*}

%% file: tables/quant_occ.tex
\newcolumntype{h}{>{\columncolor{myazure}}c}
\begin{table*}[t!]
\centering
\caption{
Quantitative results for videos affected by human-induced occlusions and the presence of distractor chairs, reported in the same format as Tab.~\ref{tab:quant_noocc}.
}
\label{tab:quant_occ}

\resizebox{\columnwidth}{!}{%
\begin{tabular}{rlchhcchh}
    \toprule
    & \multirow{2}{*}{Method} & Localization & \multicolumn{2}{c}{\cellcolor{myazure}\zed} & \multicolumn{2}{c}{\rsd} & \multicolumn{2}{c}{\cellcolor{myazure}\rsl} \\
    & & prior & $E^R$ & $E^T$ & $E^R$ & $E^T$ & $E^R$ & $E^T$ \\
    \toprule
    {\color{gray} \scriptsize 1} & \multirow{1}{*}{FreeZe-FPFH~\cite{rusu2009fpfh}} & & $130 \pm 37$ & $1850 \pm 466$ & $124 \pm 36$ & $1343 \pm 523$ & $125 \pm 42$ & $693 \pm 453$ \\
    {\color{gray} \scriptsize 2} & \multirow{1}{*}{SAM-6D~\cite{lin2024sam6d}} & & $121 \pm 35$ & $1866 \pm 362$ & $126 \pm 34$ & $4252 \pm 1088$ & $121 \pm 37$ & $665 \pm 329$   \\
    {\color{gray} \scriptsize 3} & \multirow{1}{*}{FreeZe-GeDi~\cite{poiesi2023gedi}} & & $94 \pm 64$ & $701 \pm 747$ & $107 \pm 56$ & $935 \pm 733$ & $68 \pm 61$ & $480 \pm 534$ \\
    \midrule
    {\color{gray} \scriptsize 4} & \multirow{1}{*}{FreeZe-FPFH~\cite{rusu2009fpfh}} & ZS & $ 126\pm43 $ & $ 1389\pm 716$ & $ 125\pm 41$ & $ 995\pm 1357$ & $ 123\pm 42$ & $ 902\pm 604$ \\
    {\color{gray} \scriptsize 5} & \multirow{1}{*}{SAM-6D~\cite{lin2024sam6d}} & ZS & $ 96\pm 136$ & $ 1884\pm 1419$ & $ 93\pm 49$ & $5441 \pm 5346$ & $ 104\pm 49 $ & $1263 \pm 684$\\
    {\color{gray} \scriptsize 6} & \multirow{1}{*}{FreeZe-GeDi~\cite{poiesi2023gedi}} & ZS & $ 77\pm 63$ & $ 541\pm 754$ & $ 102\pm 61$ & $ 841\pm 706$ & $ 85\pm 58$ & $ 741\pm 617$ \\
    \midrule
    {\color{gray} \scriptsize 7} & \multirow{1}{*}{FreeZe-FPFH~\cite{rusu2009fpfh}} & GT & $111 \pm 53$ & $181 \pm 155$ & $118 \pm 47$ & $178 \pm 113$ & $117 \pm 51$ & $226 \pm 199$ \\
    {\color{gray} \scriptsize 8} & \multirow{1}{*}{SAM-6D~\cite{lin2024sam6d}} & GT & $30 \pm 47$ & $158 \pm 290$ & $84 \pm 55$ & $1037 \pm 1053$ & $57 \pm 54$ & $472 \pm 474$ \\
    {\color{gray} \scriptsize 9} & \multirow{1}{*}{FreeZe-GeDi~\cite{poiesi2023gedi}} & GT & $40 \pm 57$ & $97 \pm 165$ & $75 \pm 69$ & $138 \pm 130$ & $45 \pm 51$ & $124 \pm 192$ \\
    \bottomrule
\end{tabular}
}
\end{table*}

%% file: figures/qualitative_results/qualitative_results.tex
\begin{figure}[t!]
\centering

\begin{minipage}[c]{0.02\textwidth}

\end{minipage}
\begin{minipage}[c]{0.45\textwidth}
    \centering
    \footnotesize Full chair
\end{minipage}
\hspace{1mm}
\begin{minipage}[c]{0.47\textwidth}
    \centering
    \footnotesize Frame-only chair
\end{minipage}

\begin{minipage}[c]{0.02\textwidth}
\end{minipage}
\begin{minipage}[c]{0.235\textwidth}
    \centering
    \footnotesize Entire scene
\end{minipage}
\begin{minipage}[c]{0.235\textwidth}
    \centering
    \footnotesize Zero-shot segmentation
\end{minipage}
\hspace{1mm}
\begin{minipage}[c]{0.235\textwidth}
    \centering
    \footnotesize Entire scene
\end{minipage}
\begin{minipage}[c]{0.235\textwidth}
    \centering
    \footnotesize Zero-shot segmentation
\end{minipage}

\hspace{0.02\textwidth} 
\hspace{-1mm}
\begin{minipage}[c]{0.235\textwidth}
    \centering
    \begin{overpic}[width=\linewidth]{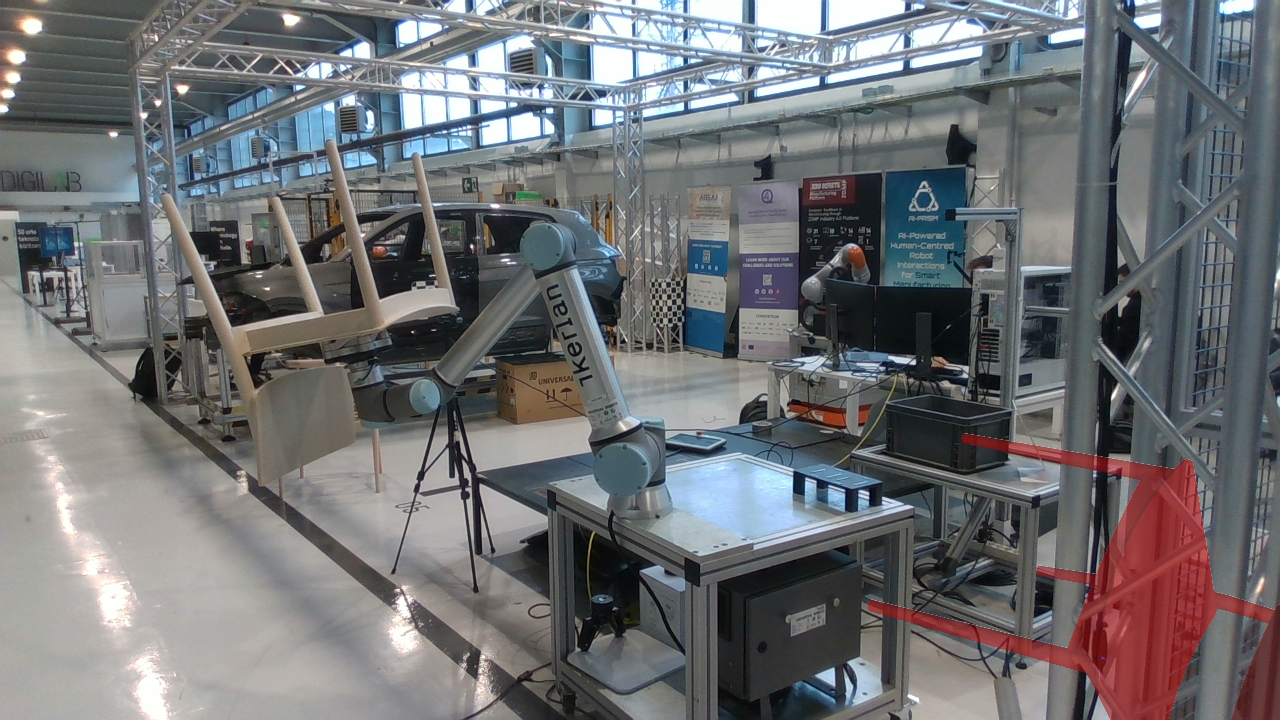} 
    \end{overpic} 
    \tiny $E^R=106.9^\circ$, $E^T=2132$mm
\end{minipage}
\begin{minipage}[c]{0.235\textwidth}
    \centering
    \begin{overpic}[width=\linewidth]{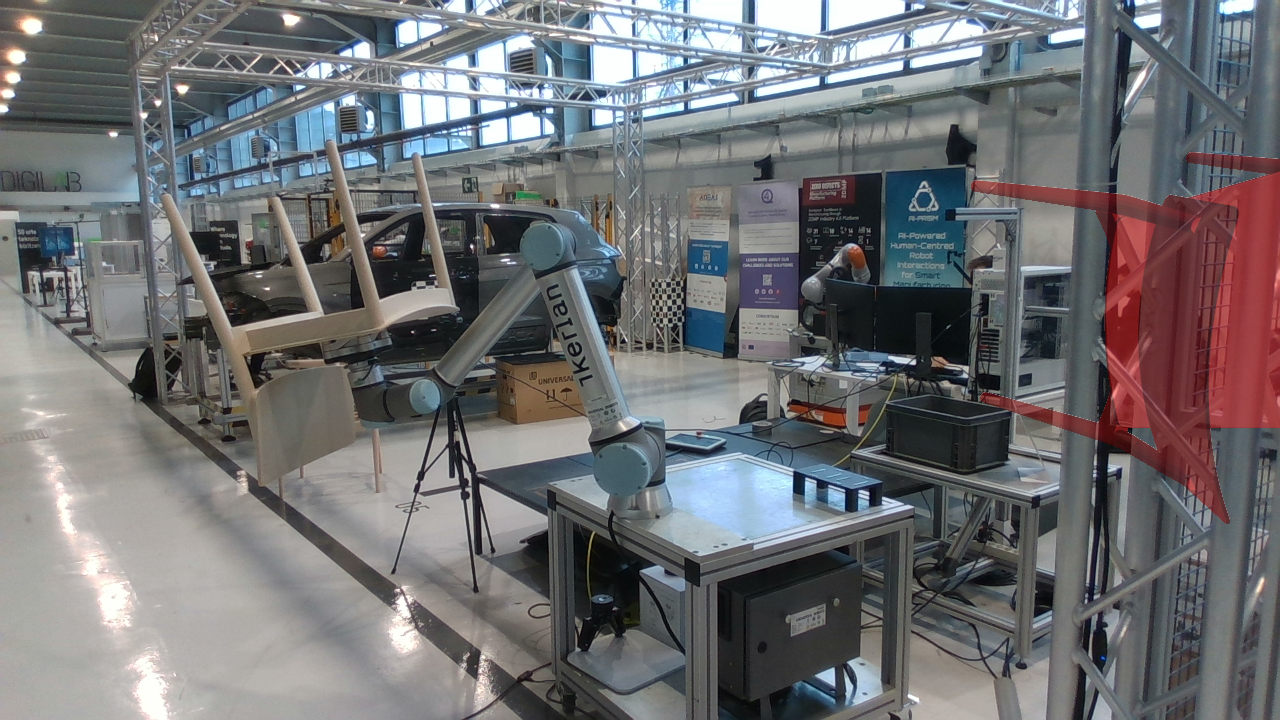} 
    \put(-3,0){
    \includegraphics[width=0.3\linewidth]{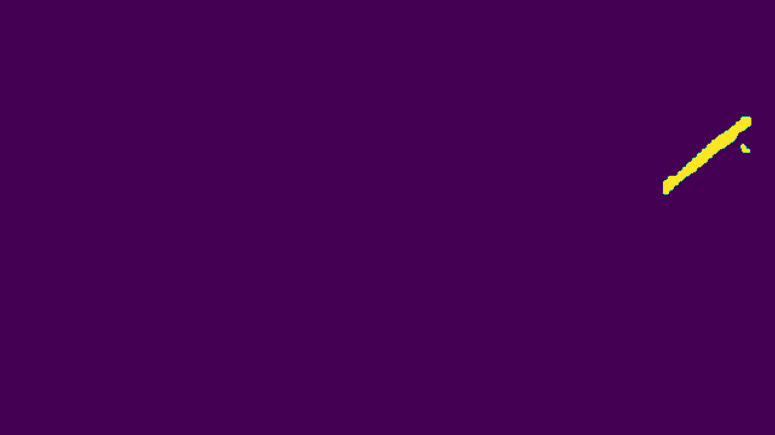}
    }
    \end{overpic} 
    
    \tiny $E^R=85.1^\circ$, $E^T=2058$mm
\end{minipage}
\hspace{1mm} 
\begin{minipage}[c]{0.235\textwidth}
    \centering
    \begin{overpic}[width=\linewidth]{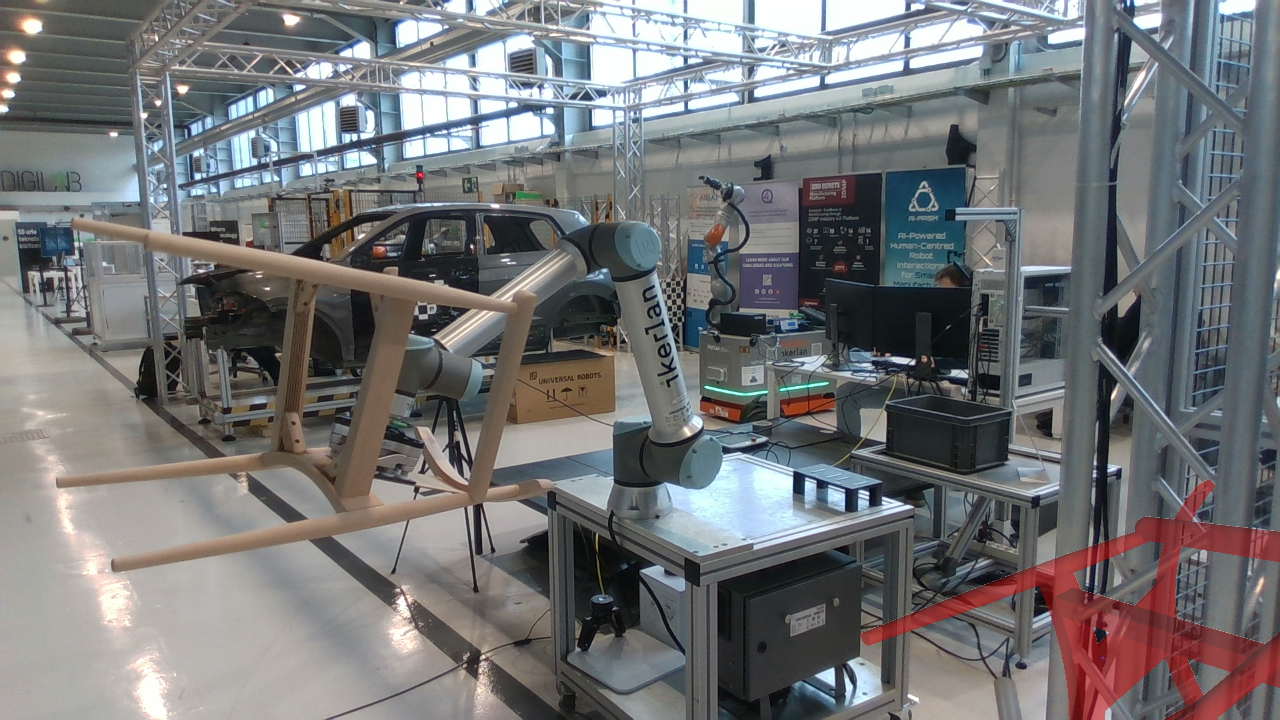} 
    \end{overpic} 
    
    \tiny $E^R=85.3^\circ$, $E^T=1594$mm
\end{minipage}
\begin{minipage}[c]{0.235\textwidth}
    \centering
    \begin{overpic}[width=\linewidth]{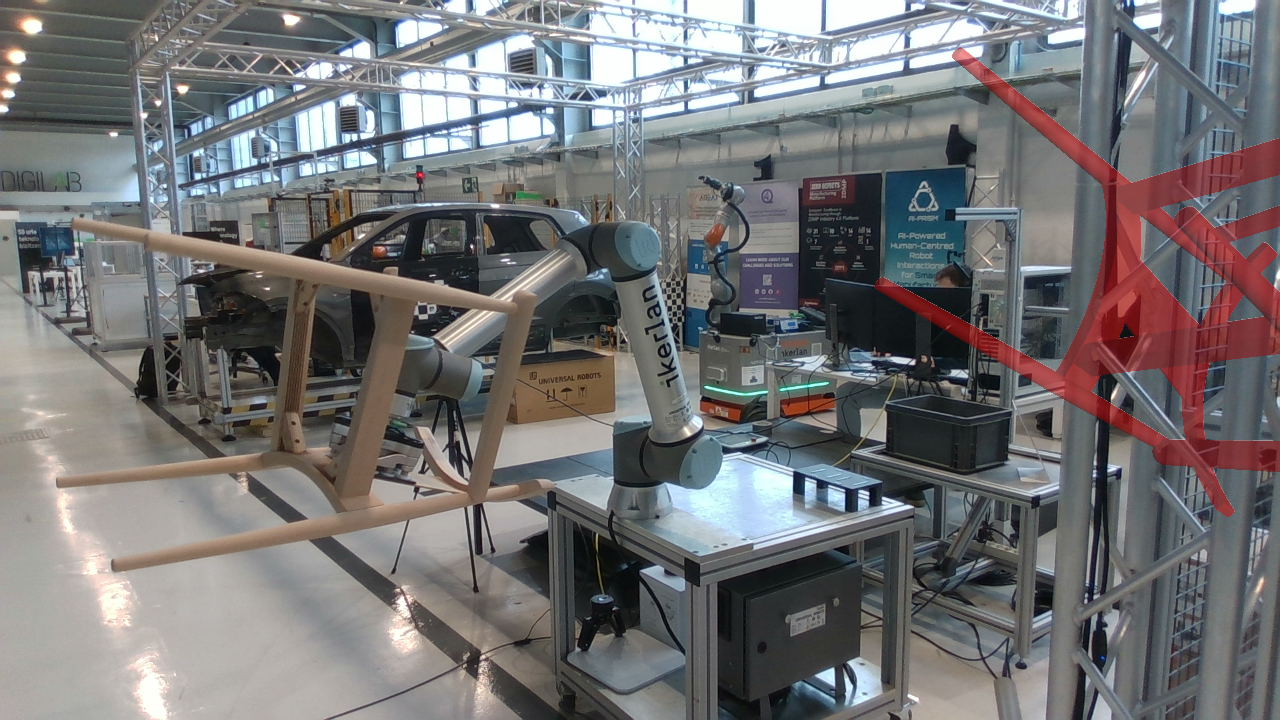} 
    \put(-3,0){
    \includegraphics[width=0.3\linewidth]{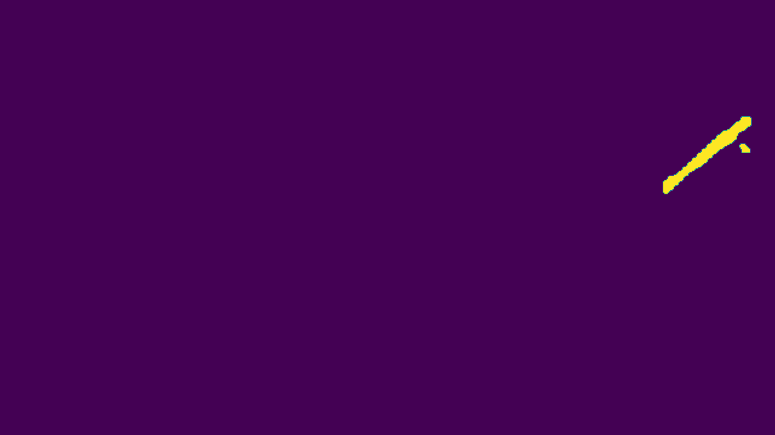}
    }
    \end{overpic} 
    
    \tiny $E^R=82.7^\circ$, $E^T=1557$mm
\end{minipage}

\vspace{1.1mm}

\begin{minipage}[c]{0.02\textwidth}
    \centering
    \vspace{-22mm} 
    \rotatebox{90}{\footnotesize No human-induced occlusions}
\end{minipage}
\begin{minipage}[c]{0.235\textwidth}
    \centering
    \begin{overpic}[width=\linewidth]{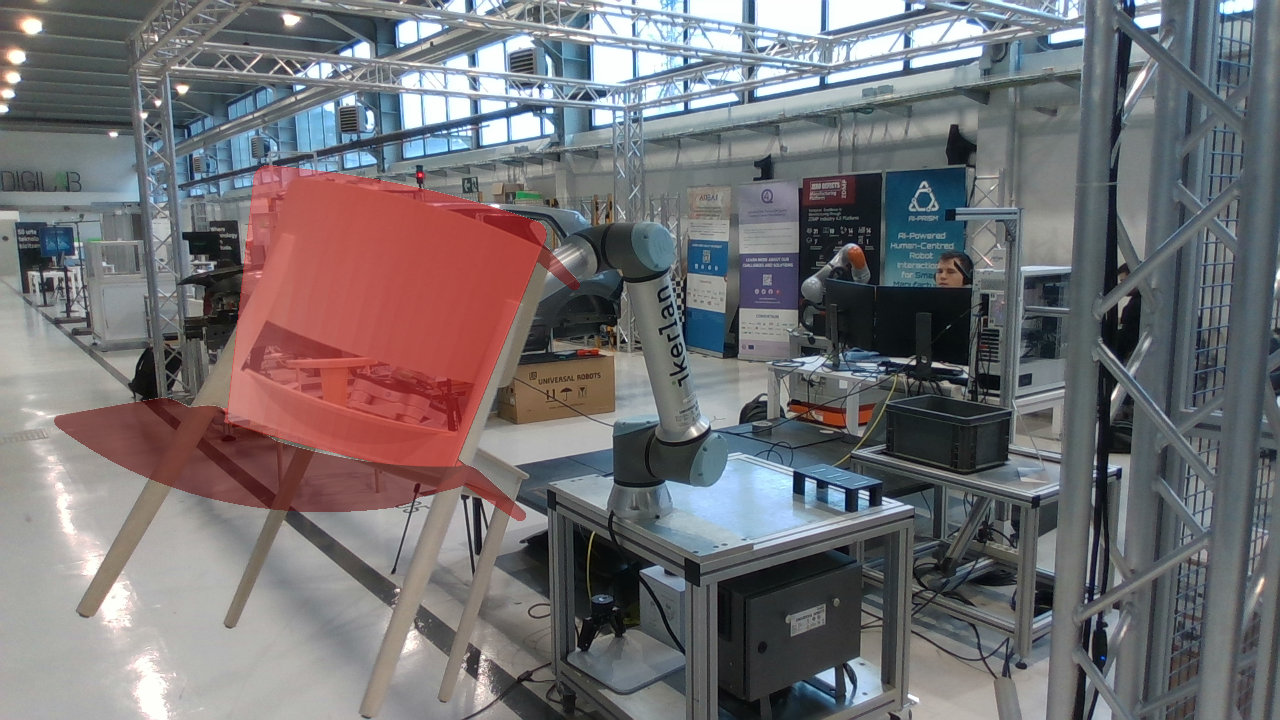}
    \end{overpic} 
    \tiny $E^R=99.0^\circ$, $E^T=190$mm
\end{minipage}
\begin{minipage}[c]{0.235\textwidth}
    \centering
    \begin{overpic}[width=\linewidth]{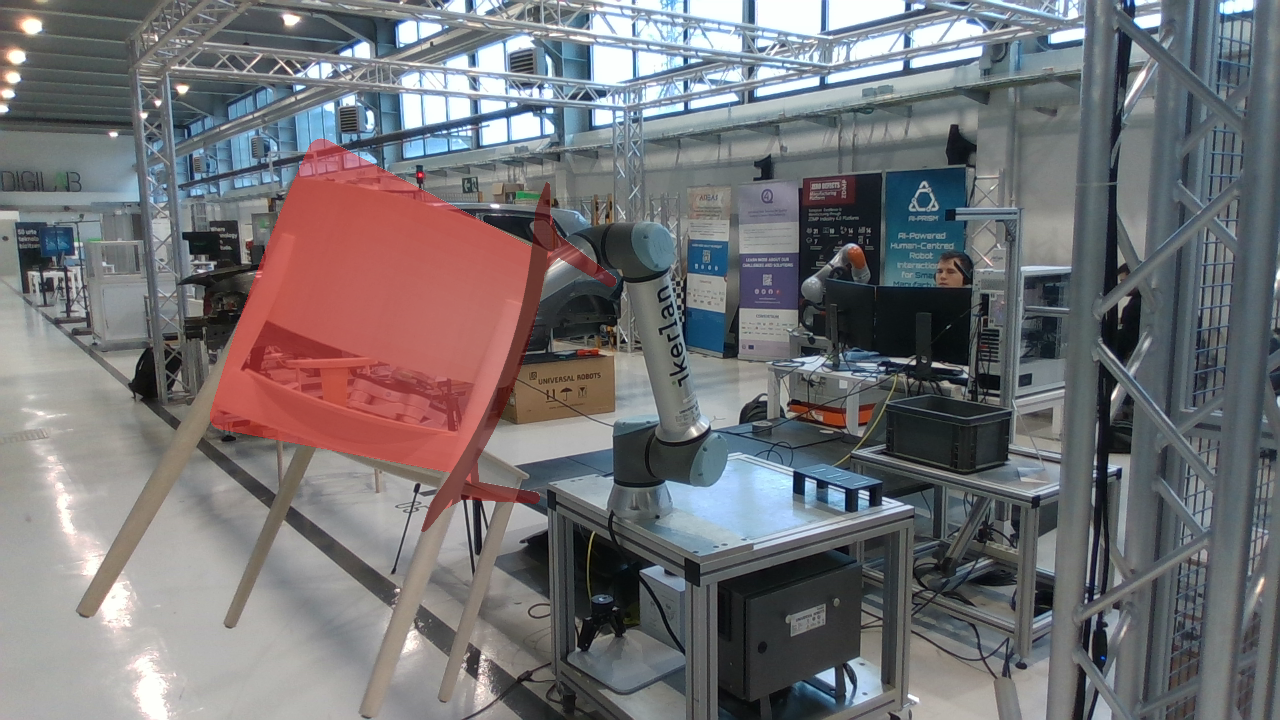}
    \put(-3,0){
    \includegraphics[width=0.3\linewidth]{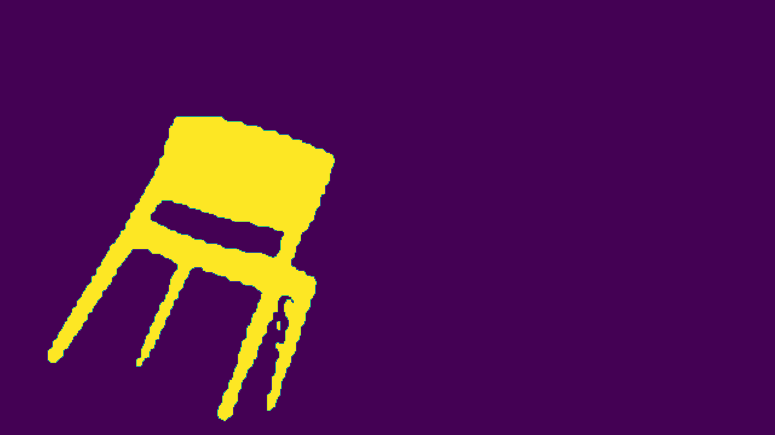}
    }
    \end{overpic} 
    \tiny $E^R=127.4^\circ$, $E^T=222$mm
\end{minipage}
\hspace{1mm} 
\begin{minipage}[c]{0.235\textwidth}
    \centering
    \begin{overpic}[width=\linewidth]{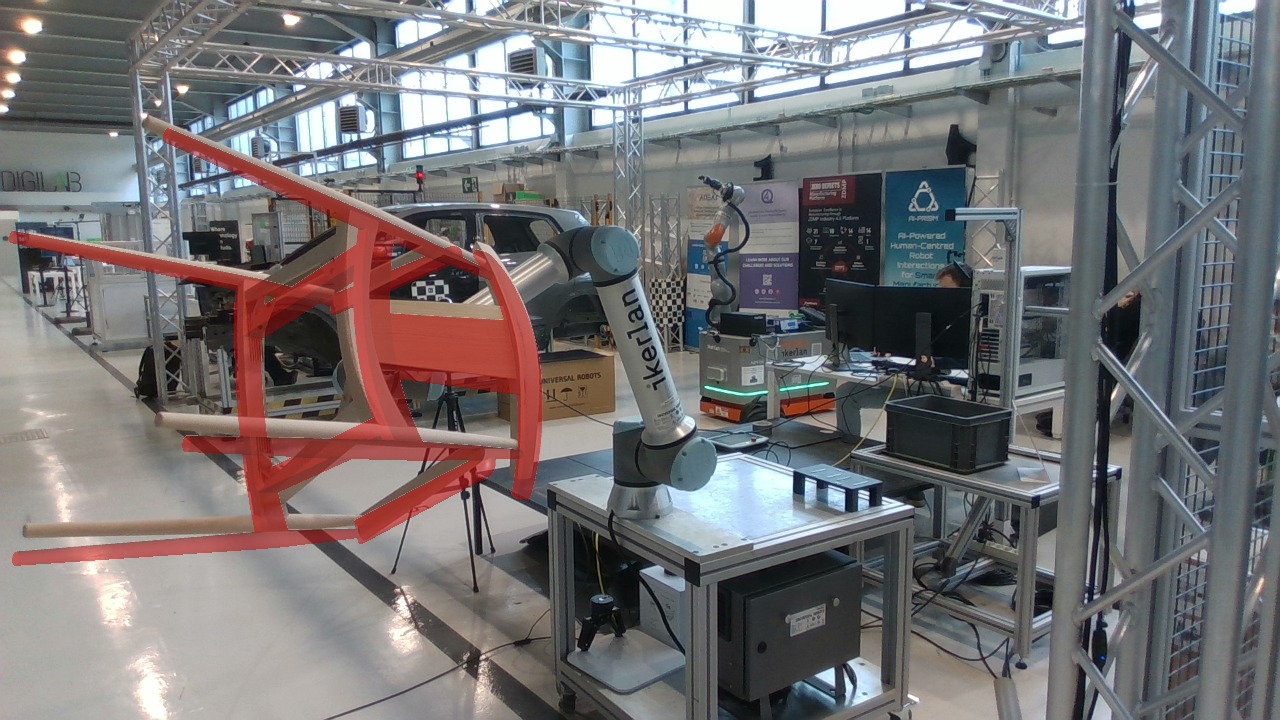} 
    \end{overpic} 
    \tiny $E^R=2.3^\circ$, $E^T=48$mm
\end{minipage} 
\begin{minipage}[c]{0.235\textwidth}
    \centering
    \begin{overpic}[width=\linewidth]{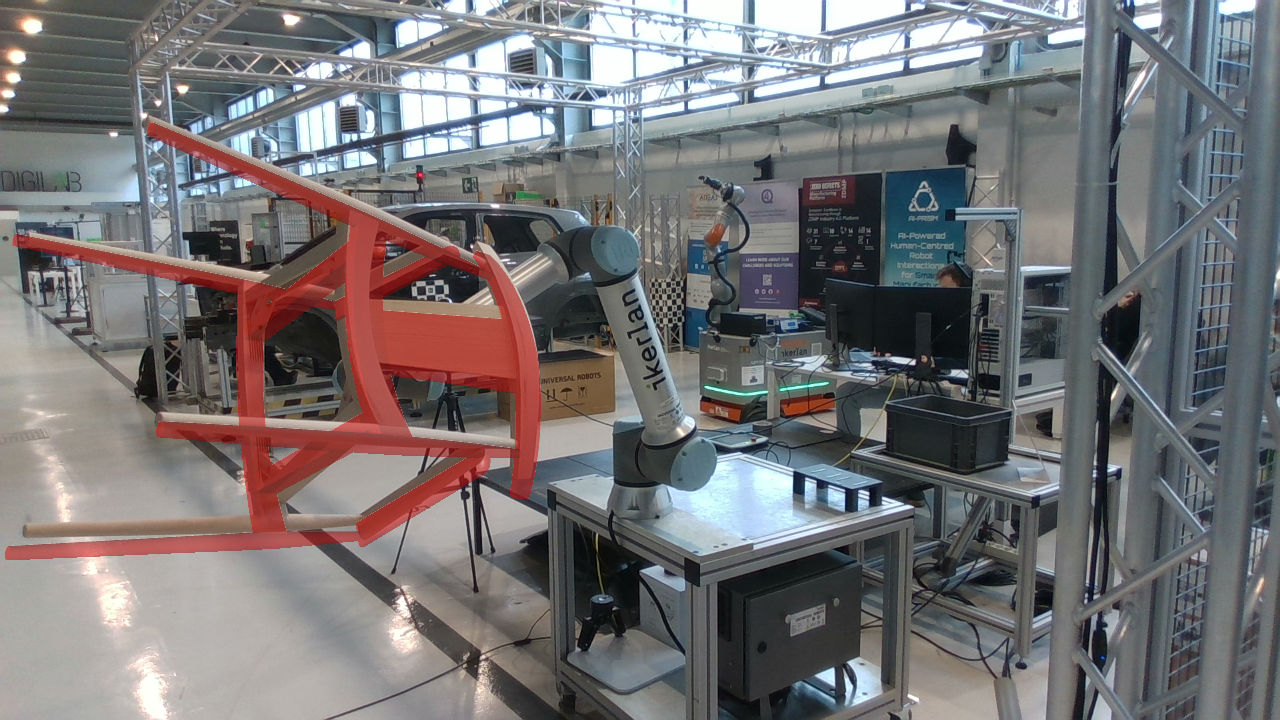} 
    \put(-3,0){
    \includegraphics[width=0.3\linewidth]{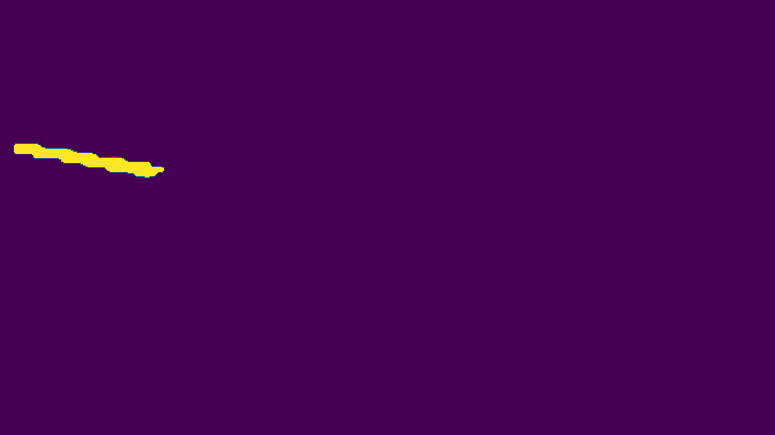}
    }
    \end{overpic} 
    \tiny $E^R=2.4^\circ$, $E^T=46$mm
\end{minipage} 

\vspace{1.7mm} 

\hspace{0.02\textwidth} 
\begin{minipage}[c]{0.235\textwidth}
    \centering
    \begin{overpic}[width=\linewidth]{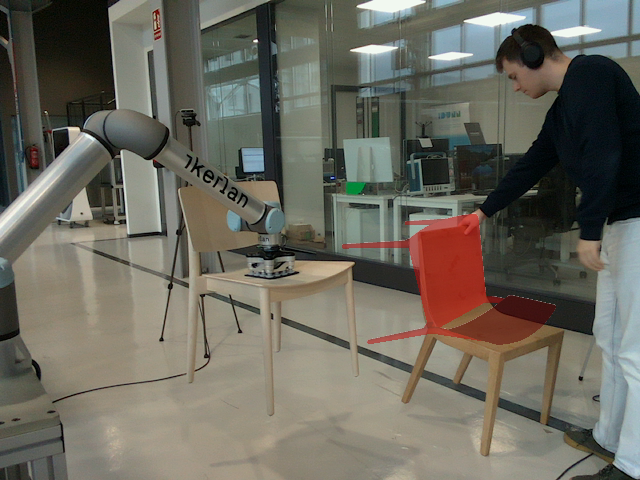}
    \end{overpic} 
    \tiny $E^R=179.9^\circ$, $E^T=951$mm
\end{minipage}
\begin{minipage}[c]{0.235\textwidth}
    \centering
    \begin{overpic}[width=\linewidth]{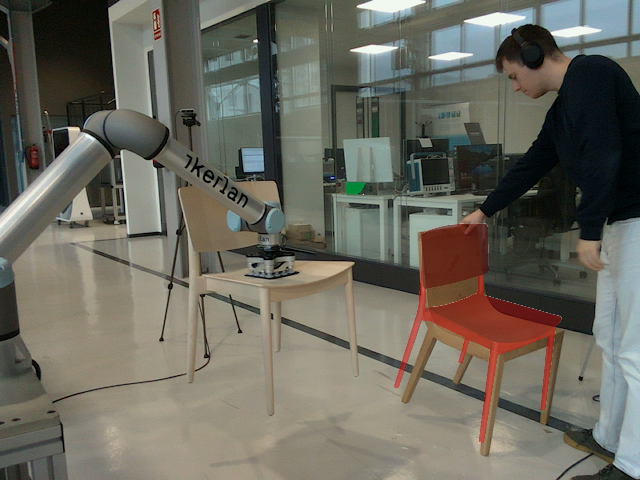}
    \put(-3,0){
    \includegraphics[width=0.3\linewidth]{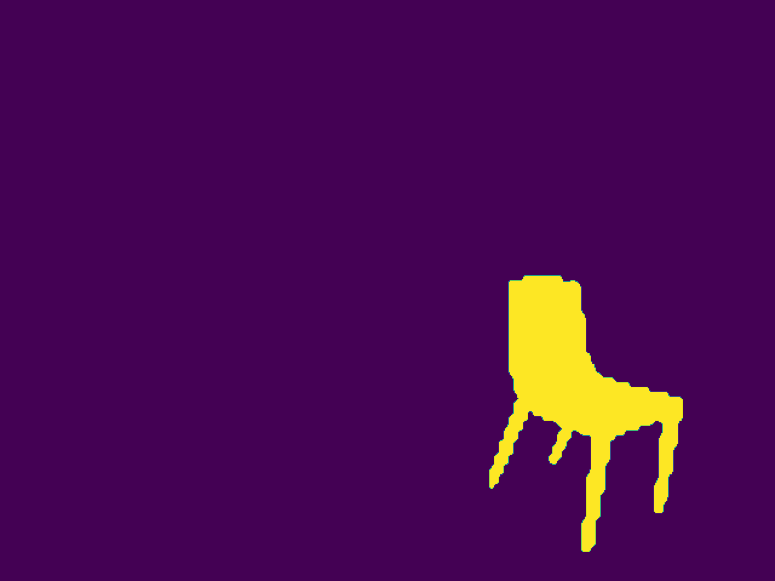}
    }
    \end{overpic} 
    \tiny $E^R=13.2^\circ$, $E^T=1033$mm
\end{minipage}
\hspace{1mm}
\begin{minipage}[c]{0.235\textwidth}
    \centering
    \begin{overpic}[width=\linewidth]{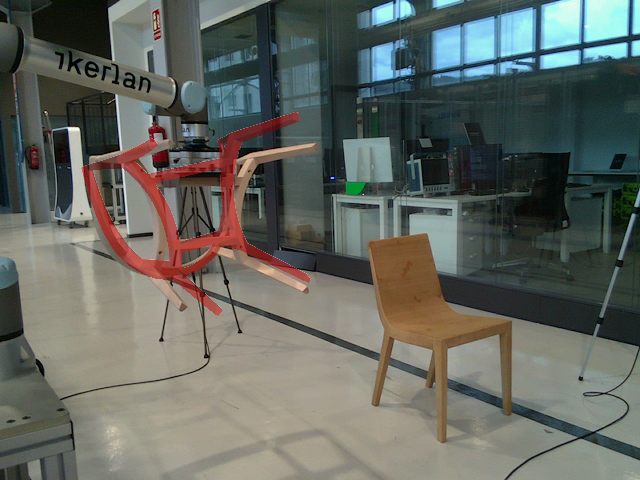}
    \end{overpic} 
    \tiny $E^R=10.9^\circ$, $E^T=34$mm
\end{minipage}
\begin{minipage}[c]{0.235\textwidth}
    \centering
    \begin{overpic}[width=\linewidth]{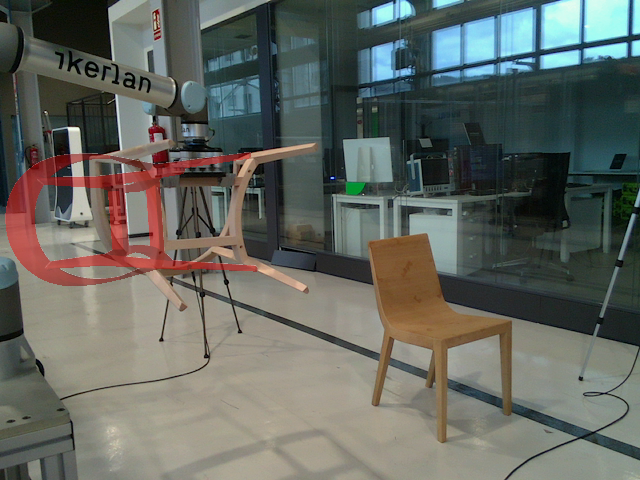}
    \put(-3,0){
    \includegraphics[width=0.3\linewidth]{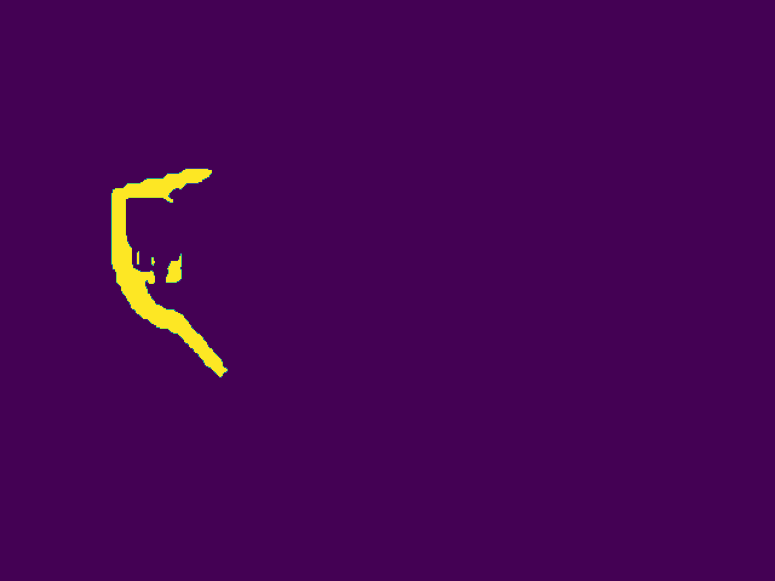}
    }
    \end{overpic} 
    \tiny $E^R=97.2^\circ$, $E^T=413$mm
\end{minipage}

\vspace{1.4mm} 

\begin{minipage}[c]{0.02\textwidth}
    \centering
    \vspace{-27mm}
    \rotatebox{90}{\footnotesize Presence of a distractor chair}
\end{minipage}
\begin{minipage}[c]{0.235\textwidth}
    \centering
    \begin{overpic}[width=\linewidth]{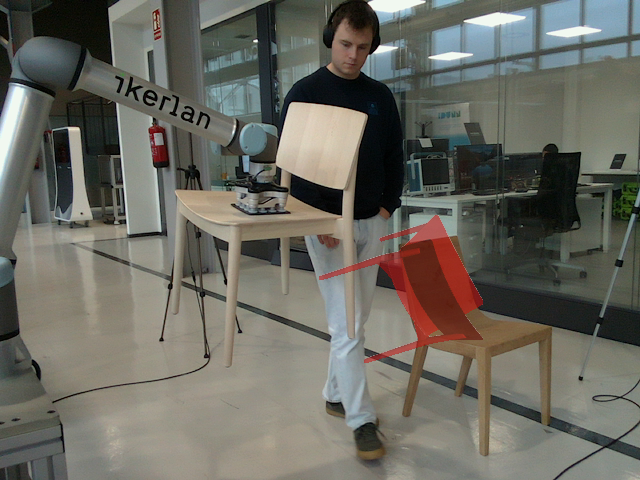}
    \end{overpic} 
    \tiny $E^R=160.5^\circ$, $E^T=1049$mm
\end{minipage}
\begin{minipage}[c]{0.235\textwidth}
    \centering
    \begin{overpic}[width=\linewidth]{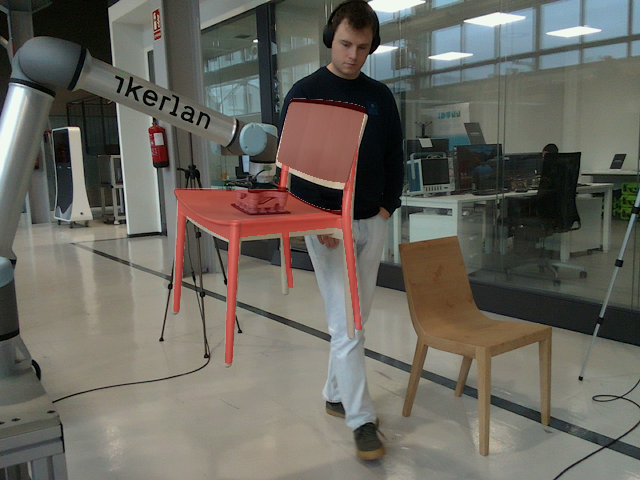}
    \put(-3,0){
    \includegraphics[width=0.3\linewidth]{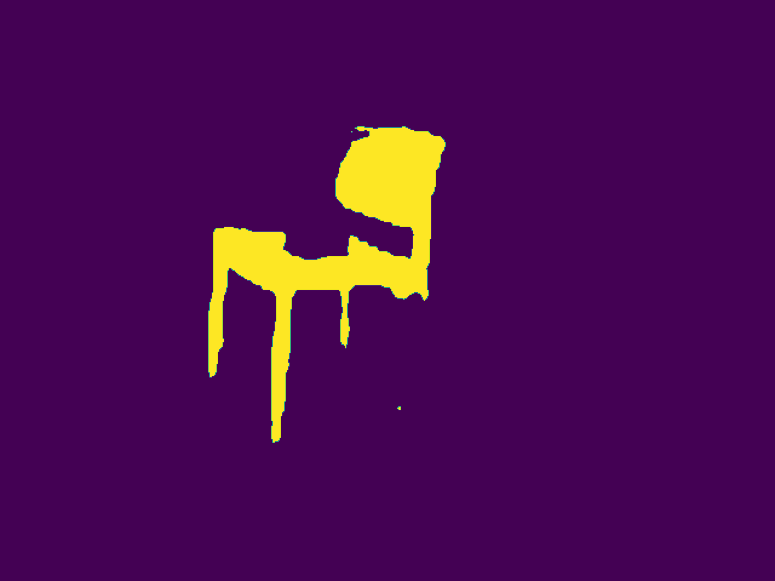}
    }
    \end{overpic} 
    \tiny $E^R=1.4^\circ$, $E^T=43$mm
\end{minipage}
\hspace{1mm}
\begin{minipage}[c]{0.235\textwidth}
    \centering
    \begin{overpic}[width=\linewidth]{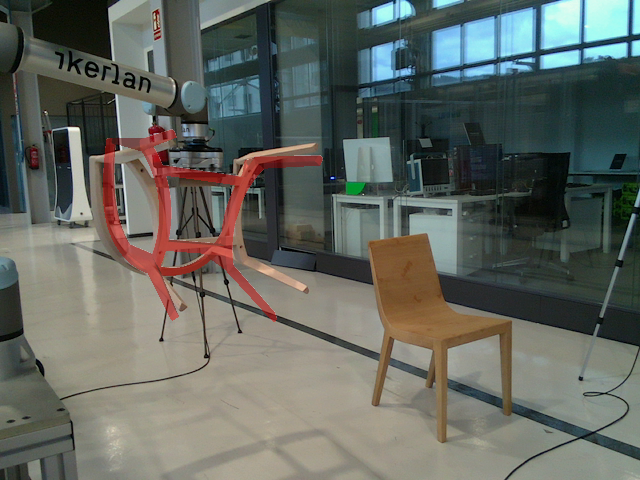}
    \end{overpic} 
    \tiny $E^R=10.5^\circ$, $E^T=37$mm
\end{minipage}
\begin{minipage}[c]{0.235\textwidth}
    \centering
    \begin{overpic}[width=\linewidth]{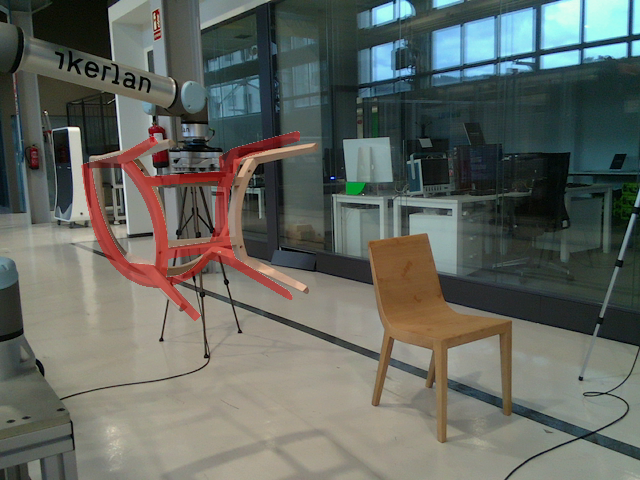}
    \put(-3,0){
    \includegraphics[width=0.3\linewidth]{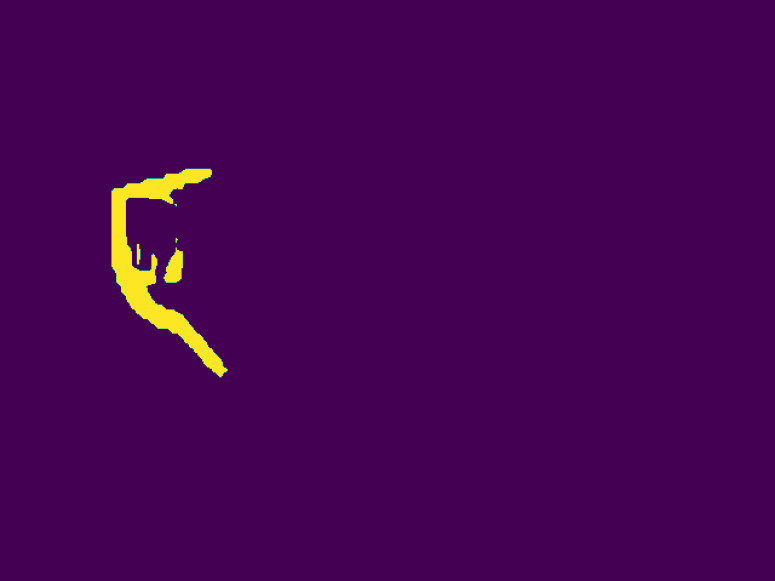}
    }
    \end{overpic} 
    \tiny $E^R=3.2^\circ$, $E^T=70$mm
\end{minipage}

\vspace{1mm} 

\hspace{0.02\textwidth} 
\begin{minipage}[c]{0.235\textwidth}
    \centering
    \begin{overpic}[width=\linewidth]{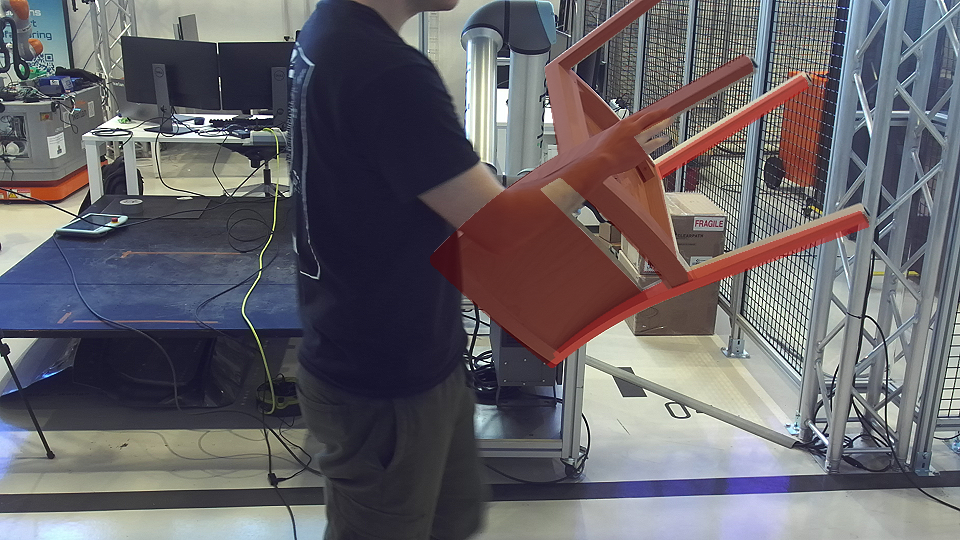}  
    \end{overpic} 
    \tiny $E^R=4.6^\circ$, $E^T=26$mm
\end{minipage}
\begin{minipage}[c]{0.235\textwidth}
    \centering
    \begin{overpic}[width=\linewidth]{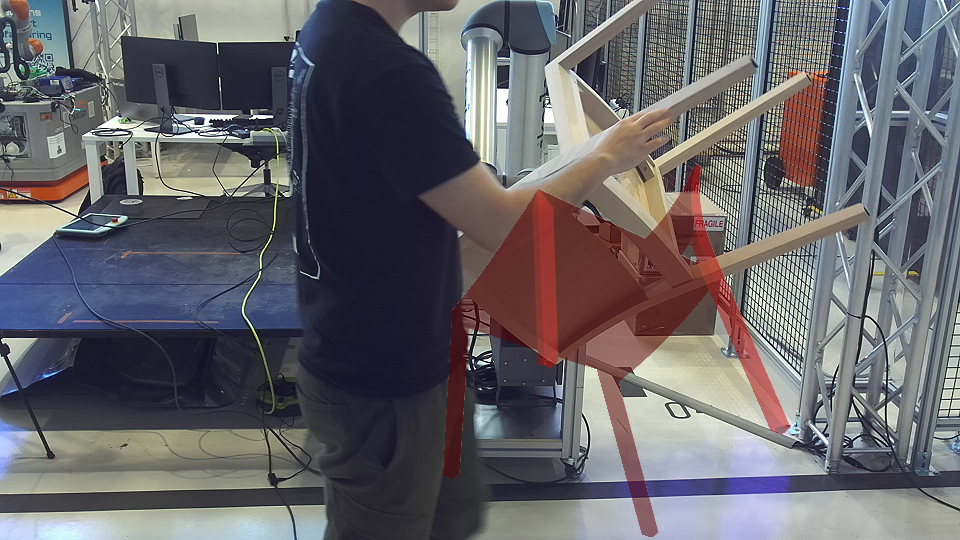} 
    \put(-3,0){
    \includegraphics[width=0.3\linewidth]{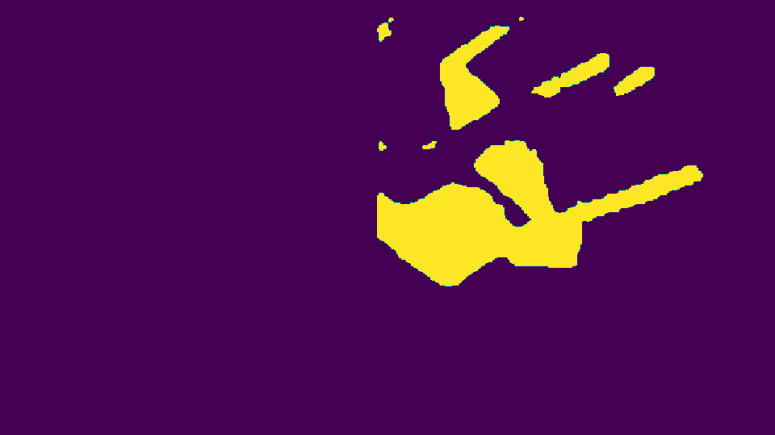}
    }
    \end{overpic} 
    \tiny $E^R=139.5^\circ$, $E^T=258$mm
\end{minipage}
\hspace{1mm}
\begin{minipage}[c]{0.235\textwidth}
    \centering
    \begin{overpic}[width=\linewidth]{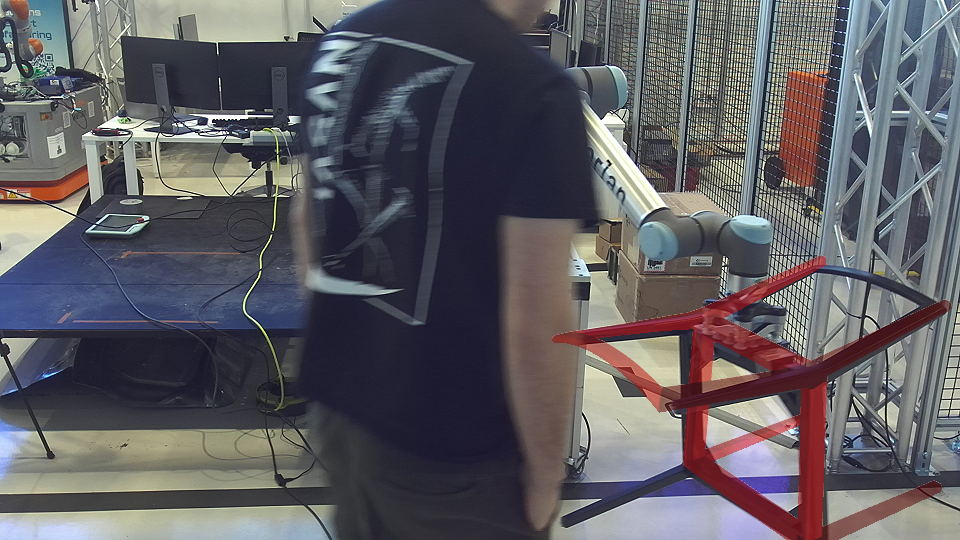} 
    \end{overpic} 
    \tiny $E^R=174.9^\circ$, $E^T=13$mm
\end{minipage}
\begin{minipage}[c]{0.235\textwidth}
    \centering
    \begin{overpic}[width=\linewidth]{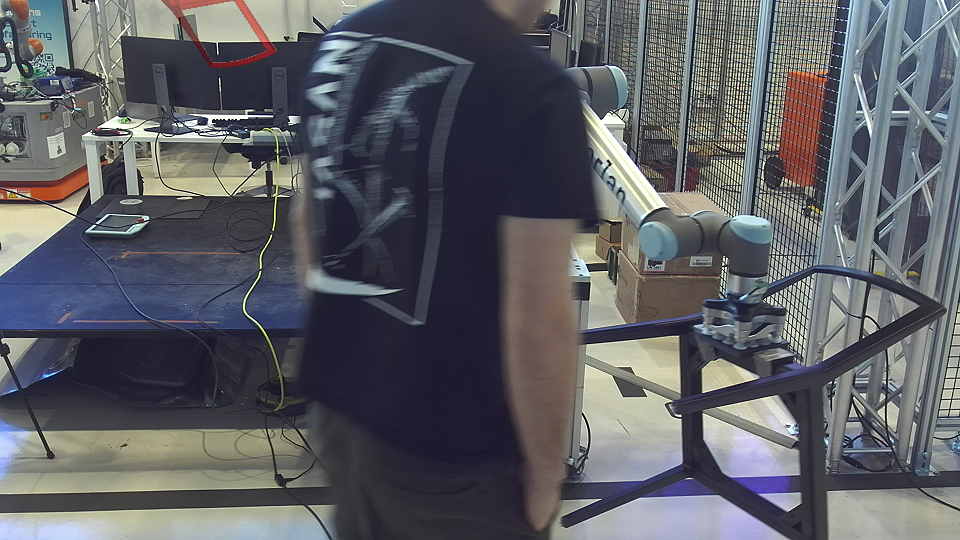}
    \put(-3,0){
    \includegraphics[width=0.3\linewidth]{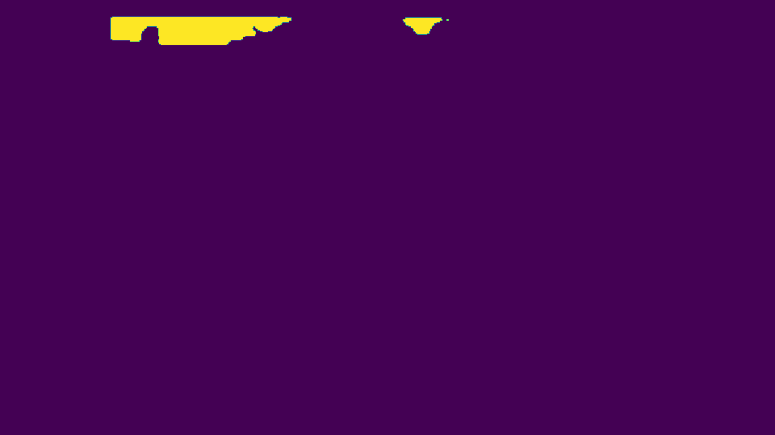}
    }
    \end{overpic} 
    \tiny $E^R=153.4^\circ$, $E^T=4037$mm
\end{minipage}

\vspace{1mm} 

\begin{minipage}[c]{0.02\textwidth}
    \centering
    \vspace{-22mm}
    \rotatebox{90}{\footnotesize Human-induced occlusions}
\end{minipage}
\begin{minipage}[c]{0.235\textwidth}
    \centering
    \begin{overpic}[width=\linewidth]{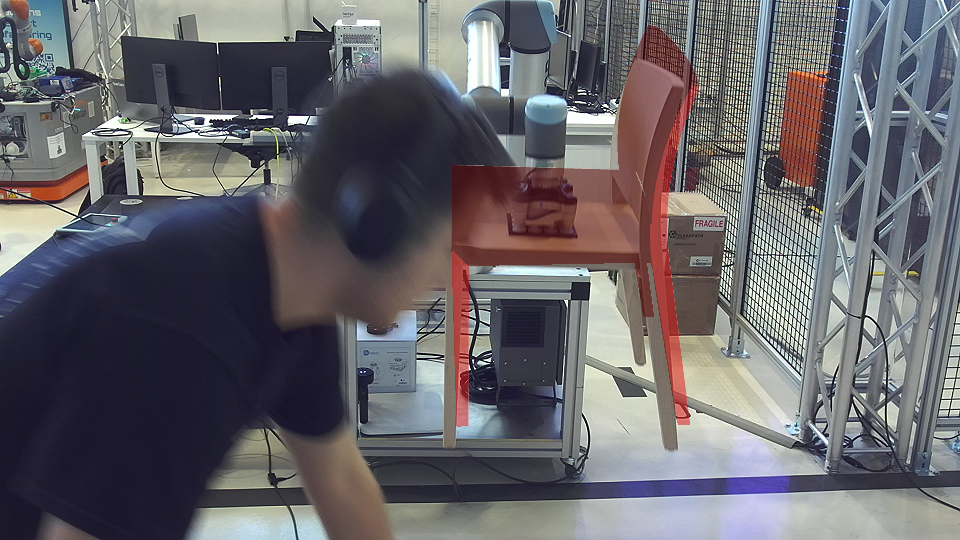}
    \end{overpic} 
    \tiny $E^R=12.4^\circ$, $E^T=84$mm
\end{minipage}
\begin{minipage}[c]{0.235\textwidth}
    \centering
    \begin{overpic}[width=\linewidth]{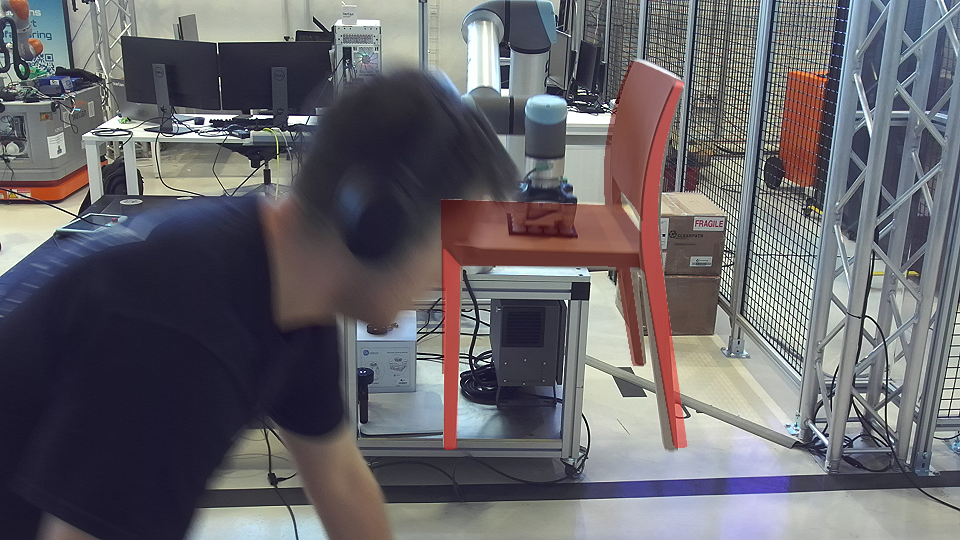}
    \put(-3,0){
    \includegraphics[width=0.3\linewidth]{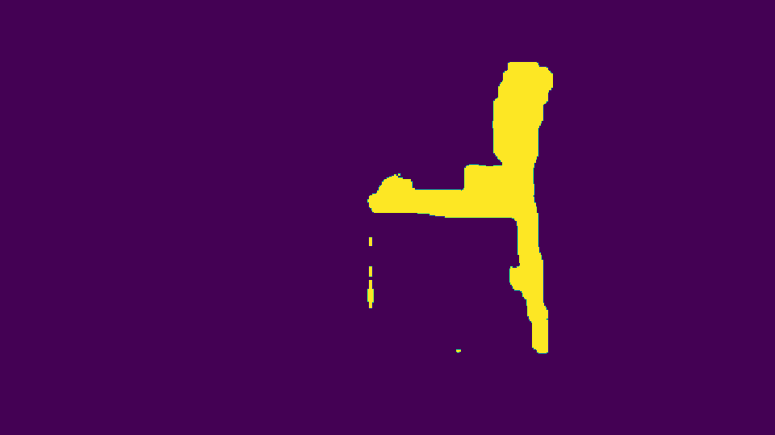}
    }
    \end{overpic} 
    \tiny $E^R=2.2^\circ$, $E^T=52$mm
\end{minipage}
\hspace{1mm}
\begin{minipage}[c]{0.235\textwidth}
    \centering
    \begin{overpic}[width=\linewidth]{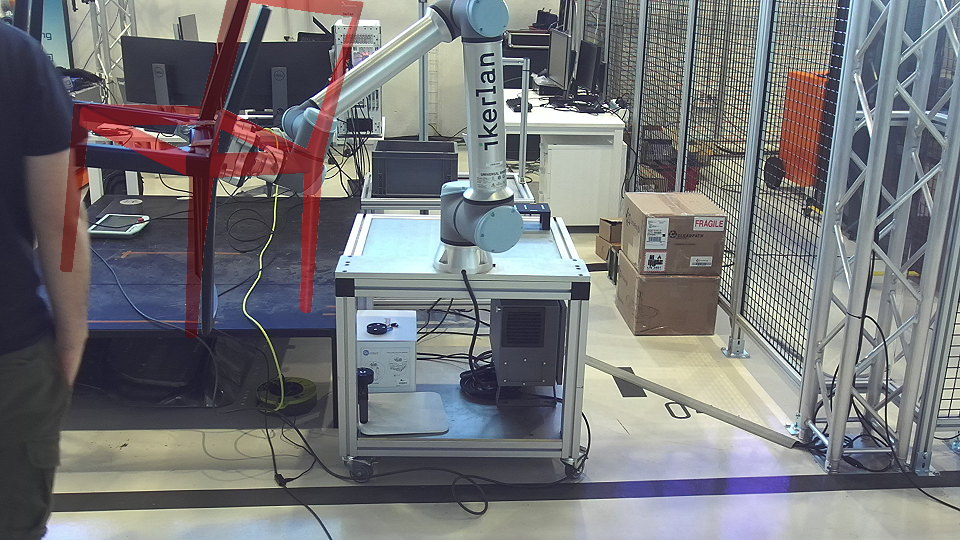}
    \end{overpic} 
    \tiny $E^R=178.0^\circ$, $E^T=201$mm
\end{minipage}
\begin{minipage}[c]{0.235\textwidth}
    \centering
    \begin{overpic}[width=\linewidth]{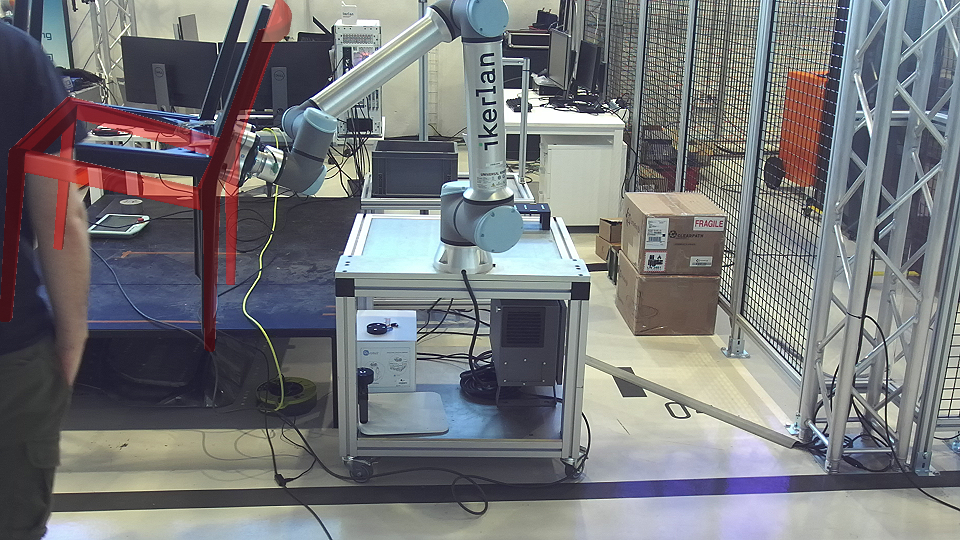}
    \put(-3,0){
    \includegraphics[width=0.3\linewidth]{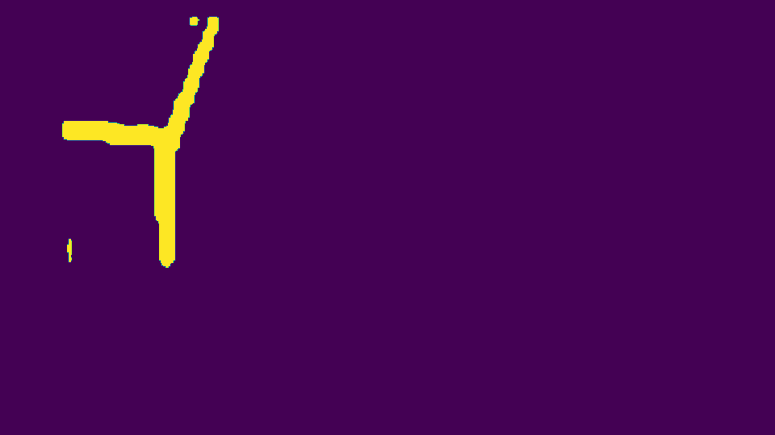}
    }
    \end{overpic} 
    \tiny $E^R=173.5^\circ$, $E^T=70$mm
\end{minipage}

\vspace{-2mm}
\caption{
Qualitative results for FreeZe-GeDi.
For each image, we overlay the CAD model transformed according to the predicted pose (shown in red for a better contrast).
Columns show different chair types (full vs. frame-only) and compare results obtained using the entire scene or only the region identified by the zero-shot segmentation (shown in the bottom left corner).
Rows show different challenges: no occlusions, presence of a distractor chair, and mild human-induced occlusion.
To facilitate quantitative comparison, we report $E^R$ and $E^T$.
}
\label{fig:qualitative_results}
\end{figure}

%% file: sections/5_conclusion.tex
\section{Conclusions}\label{sec:conclusion}

We presented \acronym, a novel 6D pose estimation dataset of wooden chairs in industrial environments.
Unlike existing datasets, \acronym employs a robotic arm to manipulate the chairs, both to simulate realistic industrial conditions and to obtain ground-truth 6D poses via the robot's kinematics.
Initial results indicate substantial room for improvement, highlighting the unique challenges it poses compared to existing benchmarks.
We will publicly release \acronym to support the evaluation of 6D pose estimation methods in realistic industrial settings.

As future work, we will extend \acronym to support object-agnostic 6D pose estimation, where the identity of the manipulated chair is undisclosed, and to enable 6D pose tracking during robotic arm movements.
We will also integrate \acronym into the recently introduced BOP-Industrial Benchmark to reach a broader audience.

\noindent\textbf{Acknowledgements.} 
This work was supported by the European Union’s Horizon Europe research and innovation programme under grant agreement No. 101058589 (AI-PRISM).